\documentclass{article}
\usepackage[T1]{fontenc}
\usepackage{iclr2027_conference,times}

\usepackage{amsmath,amsfonts,bm}

\def\eqref#1{equation~\ref{#1}}
\def\plaineqref#1{\ref{#1}}
\def\1{\bm{1}}

\DeclareMathAlphabet{\mathsfit}{\encodingdefault}{\sfdefault}{m}{sl}
\SetMathAlphabet{\mathsfit}{bold}{\encodingdefault}{\sfdefault}{bx}{n}

\usepackage{amsmath,amssymb,mathtools}
\usepackage{booktabs}
\usepackage{graphicx}
\usepackage{microtype}
\usepackage{multirow}
\usepackage{placeins}
\usepackage{float}
\usepackage{url}
\usepackage{xcolor}
\usepackage{hyperref}
\hypersetup{hidelinks}

\newfloat{algorithm}{tbp}{loa}
\floatname{algorithm}{Algorithm}

\newcommand{\CifarStandardFID}{12.421}
\newcommand{\CifarStandardKID}{.005721}
\newcommand{\CifarTimestepFID}{11.320}
\newcommand{\CifarTimestepKID}{.005004}

\newcommand{\CifarOursFID}{10.107}
\newcommand{\CifarOursKID}{.004029}

\newcommand{\CifarCombinedGain}{2.023}

\newcommand{\CifarCombinedGainCILow}{1.820}
\newcommand{\CifarCombinedGainCIHigh}{2.226}
\newcommand{\CifarCombinedKIDGain}{0.001456}

\newcommand{\CifarReplicationGain}{1.906}

\newcommand{\CifarCrossFamilyGain}{0.490}
\newcommand{\CifarCrossFamilyKIDGain}{0.000397}
\newcommand{\PublicImageNetTransferGain}{0.111}
\newcommand{\JointTHGain}{1.041}

\newcommand{\JointTHKIDGain}{0.000682}

\newcommand{\JointTHNewGain}{1.015}

\newcommand{\CifarNFEFiveFID}{11.992}
\newcommand{\CifarNFEFiveKID}{.005471}
\newcommand{\CifarContinuousFID}{10.107}
\newcommand{\CifarContinuousKID}{.004029}
\newcommand{\CifarNFESixFID}{11.591}
\newcommand{\CifarNFESixKID}{.005252}

\newcommand{\AFHQTunedGain}{0.212}

\newcommand{\AFHQTransferGain}{0.194}
\newcommand{\AFHQTransferKID}{.000098}

\newcommand{\CifarLatencyStandard}{342.09}
\newcommand{\CifarLatencyOurs}{344.93}

\newcommand{\ImageNetLatencyStandard}{212.71}
\newcommand{\ImageNetLatencyOurs}{212.54}

\newcommand{\PrefixFiveGain}{1.199}
\newcommand{\PrefixFiveLow}{1.072}
\newcommand{\PrefixFiveHigh}{1.326}
\newcommand{\PrefixUnguidedGain}{0.109}
\newcommand{\PrefixGuidedGain}{1.308}

\newcommand{\AttributionConditional}{1.479}
\newcommand{\AttributionGuidance}{-0.066}
\newcommand{\AttributionFull}{1.485}

\newcommand{\TemporalStartGain}{0.5218}
\newcommand{\TemporalJointGain}{0.1642}
\newcommand{\TemporalJointLow}{-0.7059}
\newcommand{\TemporalJointHigh}{1.0342}
\newcommand{\TemporalJointWins}{4}

\newcommand{\EndpointStartGain}{0.566}

\newcommand{\EndpointHorizonGain}{1.394}

\newcommand{\SuppFixedBTwoHGain}{0.132}

\newcommand{\SuppFixedXLTwoHGain}{0.101}

\newcommand{\SuppPrefixOfficialDID}{1.116}

\newcommand{\SuppPrefixSemidiscreteDID}{1.294}

\newcommand{\JointTwoDImages}{300,000}
\newcommand{\JointTwoDConfirmImages}{120,000}

\newcommand{\JointTwoDConfirmCells}{12}

\newcommand{\BlindFourStartMarginA}{0.2121}

\newcommand{\BlindEightFixedGainA}{0.0528}
\newcommand{\BlindFourStartMarginB}{0.2641}

\newcommand{\BlindEightFixedGainB}{0.0545}

\title{Rethinking Prediction Horizons: Decoupling Model Queries from State Updates in Few-Step MeanFlow}
\author{%
\begin{minipage}{\dimexpr\textwidth-2\tabcolsep\relax}
\centering
\bfseries
Tinglan Wang$^{1,\dagger}$,
Haixu Liu$^{1,\dagger}$,
Shaobo Chen$^{1}$,
Nuo Qian$^{1}$,\\
Qi Gao$^{1,*}$,
Kai Wen$^{1,*}$,
Yu Yao$^{2,*}$\\[0.5em]
\normalfont
$^{1}$Beijing QBoson Quantum Technology Co., Ltd.\\
$^{2}$University of Sydney\\[0.5em]
\small
$^{\dagger}$ Equal contribution.\\
$^{*}$ Corresponding authors:\\
\texttt{gaoq@boseq.com},
\texttt{wenk@boseq.com},\\
\texttt{yu.yao@sydney.edu.au}
\end{minipage}%
}
\iclrfinalcopy
\begin{document}
\maketitle

\begin{abstract}
MeanFlow predicts interval-average velocities, but conventional few-step sampling ties the interval queried from the model to the duration of the state update. We separate these two quantities at inference, fixing the model, update grid, and number of queries while varying the prediction horizon. Our central finding is that the horizon response depends on both query position and the states entering that query. Across seven independently trained CIFAR-10 checkpoints at classifier-free guidance (CFG) scale 2, a longer horizon worsens FID at early steps but improves it at the midpoint. To probe whether query time alone determines this response, we fix the midpoint query and current/subsequent guidance while changing only earlier CFG. In held-out factorial outcomes on five previously studied checkpoints, the mean midpoint horizon benefit increases from 0.109 to 1.308 FID, indicating sensitivity to the incoming-state distribution. A frozen four-query rule reduces mean CIFAR-10 FID by 2.023 without retraining or extra queries, including gains on five checkpoints trained after rule selection. A milder rule also improves mean FID by 1.041 on a stronger frozen grid, with fixed-rule FID gains on the evaluated public ImageNet-256 and AFHQ checkpoints. However, matched-budget calibration favors query-start shifting in historical CIFAR tests, and the tested Horizon allocation adds no FID benefit beyond Grid+Start on two public ImageNet models. Thus, prediction horizon is a distinct, position- and state-sensitive inference control, but not a universally preferred one.
\end{abstract}

\section{Introduction}

MeanFlow~\citep{geng2025meanflow} predicts an average velocity over an interval and conventionally uses the same interval to advance the state. This couples two decisions: how far the model predicts and how far the sampler moves. Matching them gives the intended displacement for an exact interval-average predictor; a learned predictor's approximation also depends on the requested interval. Varying prediction horizon independently of update duration exposes an inference choice hidden by their coupling.

Let $h_k$ denote the prediction horizon and $\delta_k=t_{k+1}-t_k$ the update duration. We study
\begin{equation}
x_{k+1}=x_k+\delta_k u_\theta(x_k,t_k,h_k),
\label{eq:decoupled-update}
\end{equation}
with the model, query start, state grid and query count fixed. Standard inference sets $h_k=\delta_k$; varying only $h_k$ changes the prediction interval while preserving update duration. Earlier two-time denoisers already separate prediction intervals from output coefficients~\citep{lee2026language,zilberstein2026denoisers}. Our contribution is a controlled characterization of what this choice does in pretrained few-step MeanFlow.

\begin{figure}[t]
\centering
\includegraphics[width=\linewidth]{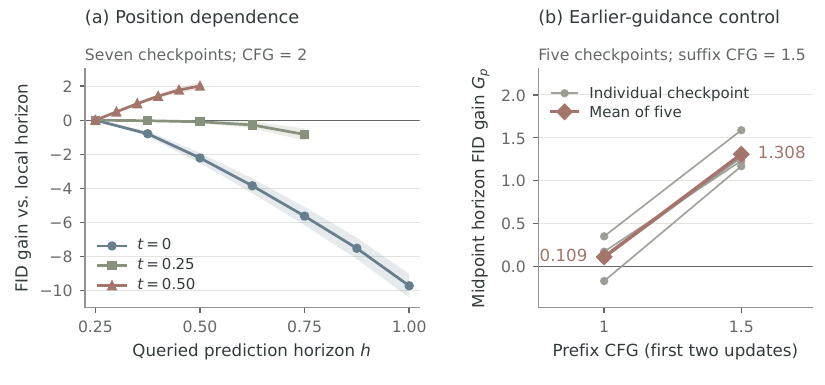}
\caption{\textbf{Horizon response reverses with position and changes with earlier guidance.} Left: seven CIFAR checkpoints at CFG 2, with pointwise 95\% Student-$t$ intervals across checkpoints. Right: five checkpoints with held-out factorial outcomes compare midpoint $h_2=.50$ versus $.25$ after prefix CFG 1 or 1.5; current/later CFG stays 1.5. Lines connect the same checkpoint. Both panels use paired 10k samples per cell, but different guidance settings and checkpoint subsets.}
\label{fig:time-horizon}
\end{figure}

The main empirical result is a \emph{position-dependent sign reversal}
in the effect of the queried horizon (Figure~1a).
With the checkpoint, state grid, update durations, and guidance fixed,
extending a single horizon from $0.25$ to $0.50$ worsens endpoint FID
at the source ($t=0$), but improves it at the midpoint ($t=0.50$)
on all seven independently trained CIFAR checkpoints at CFG~2.
A longer prediction horizon is therefore not uniformly beneficial,
even at the same extension ratio $h/\delta=2$.
This reversal exposes behavior that cannot be captured by a
global horizon multiplier.

Crucially, query position alone is insufficient to characterize
the horizon response (Figure~1b).
We hold the midpoint query time, physical grid, and
current/subsequent guidance fixed while changing only CFG in
the preceding updates.
Within each prefix condition, the two horizon choices share the
same incoming states; across prefix conditions, earlier guidance
changes the distribution of states reaching the midpoint.
In held-out factorial outcomes on five previously studied checkpoints,
with the hypothesis frozen before outcome inspection,
the mean benefit of midpoint horizon extension increases
from $0.109$ to $1.308$ FID.
This interaction establishes sensitivity to the incoming-state
distribution beyond the query time alone, although it does not
isolate the contributions of CFG construction bias,
prediction error, or other state-dependent effects.

These observations also have practical consequences at fixed
inference cost.
A frozen four-query horizon rule reduces mean CIFAR-10 FID
by $2.023$ across seven checkpoints, including five trained
after rule selection.
A milder fixed rule retains a mean gain of $1.041$ on a
stronger frozen grid, and fixed-rule FID gains extend to the
evaluated public ImageNet and AFHQ checkpoints.
However, the usefulness of horizon calibration depends on
the alternatives: query-start shifts can achieve lower FID/KID,
and the tested Horizon allocation provides no additional FID
benefit beyond Grid+Start on two public ImageNet models.
We therefore distinguish the reproducible response to prediction
horizon from whether allocating calibration budget to this
control is optimal for a given configuration and evaluator.

Our contributions are:
\begin{itemize}
    \item \textbf{Position-dependent horizon response.}
    We identify a reproducible sign reversal in the effect
    of extending the queried prediction horizon while
    preserving the physical state-update grid and
    model-query count.

    \item \textbf{Incoming-state dependence.}
    Through a preceding-guidance intervention with the
    midpoint query and subsequent guidance fixed,
    we show that the horizon benefit changes with the
    states entering the query, rather than being
    determined by query time alone.

    \item \textbf{Reproducibility and competing controls.}
    We validate frozen horizon rules across training seeds
    and a stronger state grid, assess fixed-rule transfer
    to public checkpoints, and compare horizon,
    query-start, and grid controls under explicit
    calibration budgets, identifying both benefits
    and limitations.
\end{itemize}

\section{Prediction Horizon as a Separate Control}

\subsection{Interval prediction and state evolution}

Flow Matching learns an instantaneous field $v(x,t)$ whose ODE transports a source distribution to a data distribution~\citep{lipman2023flow}. For a trajectory $X_t$, the interval-average velocity is
\begin{equation}
\bar v_h(X_t,t)=\frac{1}{h}\int_0^h v(X_{t+s},t+s)\,\mathrm{d}s,
\label{eq:average-velocity}
\end{equation}
with the instantaneous limit at $h=0$. MeanFlow learns an approximation $u_\theta(x,t,h)$ to this interval-dependent quantity~\citep{geng2025meanflow}. Standard sampling on a grid $T=(t_0,\ldots,t_K)$ uses the prediction for $h_k=\delta_k$. Changing the grid then changes both the update duration and the interval requested from the network.

Horizon control keeps $T$ and $\delta_k$ fixed. For $h>0$, write the predicted endpoint as $\widehat\Phi_h(x,t)=x+h\,u_\theta(x,t,h)$. Equation~\plaineqref{eq:decoupled-update} becomes
\begin{equation}
x_{k+1}=x_k+\frac{\delta_k}{h_k}\bigl[\widehat\Phi_{h_k}(x_k,t_k)-x_k\bigr].
\label{eq:partial-displacement}
\end{equation}
When $h_k>\delta_k$, the sampler takes a fraction of a longer-interval displacement prediction. Positive-duration queries satisfy $0\leq t_k<t_k+h_k\leq1$; $h=0$ denotes the instantaneous branch where supported.

At the first query where the horizon differs, Local and Horizon share the same entering state, query start and update duration; only the requested prediction interval changes. Their later trajectories may diverge because this query changes the predicted velocity. Query-start and state-grid controls perturb different temporal coordinates, motivating separate endpoint comparisons rather than assuming independent gains.

\begin{figure}[t]
\centering
\includegraphics[width=.88\linewidth]{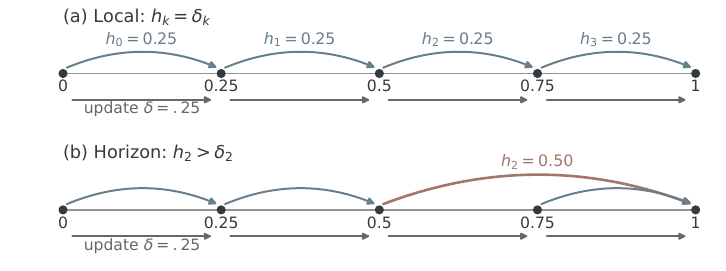}
\caption{\textbf{The prediction horizon can exceed the update duration.} Curved arrows show prediction intervals; gray arrows show physical updates. Only the midpoint horizon changes from $.25$ to $.50$. Both samplers use four updates of duration $.25$ and four velocity queries (eight backbone forwards with two-pass CFG).}
\label{fig:overview}
\end{figure}

Figure~\ref{fig:overview} makes the distinction concrete: the midpoint query can predict through time 1 while the update still stops at $.75$.

\subsection{A minimal frozen rule}

The four-query CIFAR instance uses
\begin{equation}
T=(0,.25,.50,.75,1),\quad H_{\mathrm{local}}=(.25,.25,.25,.25),\quad
H_{\mathrm{look}}=(.25,.25,.50,.25).
\label{eq:nfe4-schedule}
\end{equation}
Writing $\rho_k=h_k/\delta_k$, the rule is $\rho=(1,1,2,1)$. Selected on a historical development checkpoint and frozen before confirmation, this schedule is shared by all samples. Each update directly queries its chosen horizon, so there is no additional evaluation, per-sample selection, or retraining. A milder normalized rule, $\rho=(1,1,1.2,1)$, is used for the frozen-grid and transfer tests. Appendix~\ref{app:minimal-sampler} gives the implementation and Appendix~\ref{app:discovery} records their development.

The ratio $\rho_k=h_k/\delta_k$ preserves a relative rather than absolute horizon. Feasibility requires $\rho_k\delta_k\leq1-t_k$, making any $\rho>1$ extension infeasible at the midpoint of a uniform two-query grid. Fixed-rule transfer applies the inherited ratio only where legal; target calibration adapts to the available grid.

\subsection{Horizon mismatch and prediction error}

For a trajectory starting at $(x_t,t)$, write $\bar v_h(t)=\bar v_h(x_t,t)$. Let $\epsilon_\theta(x_t,t,h)=u_\theta(x_t,t,h)-\bar v_h(t)$ denote prediction error relative to the requested interval. Relative to the average needed for the local update,
\begin{equation}
u_\theta(x_t,t,h)-\bar v_\delta(t)
=\underbrace{\bar v_h(t)-\bar v_\delta(t)}_{\text{horizon mismatch}}
+\underbrace{\epsilon_\theta(x_t,t,h)}_{\text{interval-dependent prediction error}}.
\label{eq:error-decomposition}
\end{equation}
For a smooth trajectory, the mismatch begins with $(h-\delta)\dot v_t/2$, with the derivative taken along the trajectory (Appendix~\ref{app:derivations}). For an exact interval-average predictor, $h=\delta$ recovers the intended flow displacement over the update. For our effective predictor, $\epsilon_\theta$ can also include external CFG construction bias: mixing exact conditional and unconditional interval averages generally does not give the interval average of the guided instantaneous field. Changing $h$ changes both terms, so the horizon selects an approximation as well as a prediction interval. Endpoint metrics capture their combined effect along the resulting trajectory; the decomposition supplies a framework for interpreting this response.

\section{Reproducible Gains at Fixed Query Cost}
\label{sec:efficacy}

We first test frozen horizon rules across training seeds, on a stronger fixed grid and on other evaluated models.

\subsection{Evaluation setup}

We evaluate class-conditional CIFAR-10, public ImageNet-256 MeanFlow-family checkpoints, and latent AFHQv2 models producing $256\times256$ images~\citep{choi2020starganv2,peebles2023dit}, using FID~\citep{heusel2017fid}, KID where available~\citep{binkowski2018kid}, and specified supplementary evaluators. NFE counts velocity queries; two-pass CFG uses two backbone calls per query. Primary CIFAR tests use CFG 2, while temporal calibration uses CFG 1.5 under separate sample/reference protocols.

The primary replication unit is an independently trained checkpoint. We evaluate seven CIFAR checkpoints from one training recipe, including five trained after the horizon rule was frozen. Generation repeats are averaged within checkpoint before equal checkpoint weighting. Public ImageNet and AFHQ checkpoints separately assess fixed-rule transfer; Section~\ref{sec:blind-nfe} adds two new CIFAR seeds for calibration tests. Training definitions, checkpoint provenance and repeat accounting are in Appendices~\ref{app:training-recipes} and~\ref{app:protocol}.

\subsection{Horizon gains persist on a stronger state grid}

The frozen four-query rule improves both FID and KID on all seven CIFAR checkpoints, including all five prospectively added checkpoints. The all-seven mean FID reduction is $\CifarCombinedGain$ with a descriptive 95\% Student-$t$ interval across checkpoints $[\CifarCombinedGainCILow,\CifarCombinedGainCIHigh]$; the five-model subgroup retains $\CifarReplicationGain$. Table~\ref{tab:narrative-primary} separates the prospective subgroup from the all-seven summary; Appendix Table~\ref{tab:cifar-main} adds the original head-to-head comparison. In the original two models, Horizon also attains lower FID than Local at NFE 5 and 6 while using four velocity evaluations.

The horizon benefit also persists on a stronger state grid. We select $T^\star=(0,.3,.6,.9,1)$ from 84 predefined candidates and freeze it after an independent recheck. Changing only the third horizon from $.30$ to $.36$ on this grid yields an additional mean FID reduction of $\JointTHGain$, with FID and KID wins on 7/7 checkpoints. The five added checkpoints retain $\JointTHNewGain$.

This control tests whether the original gain merely compensates for a weak uniform grid. After selecting and freezing the stronger grid, we change only the third horizon. The remaining gain establishes a useful horizon effect on this stronger baseline. Because the two grids use different update durations and horizon ratios, their gain magnitudes are not directly comparable.

% Editorial split of unchanged sections/reviewed_primary_rows_20260923.tex.
% Caption absolute values retain panel A of sections/reviewed_controls_20260924.tex.
% Sources and exact displayed values: evidence/narrative_20260926/table_provenance.json.
\begin{table}[H]
\centering
\small
\setlength{\tabcolsep}{4pt}
\caption{\textbf{Horizon gains persist on a stronger frozen state grid.}
FID gains (Local minus Horizon) and FID/KID wins at CFG 2, four queries,
and 50k generated / 50k real images.
Means weight checkpoints equally; brackets are descriptive 95\% Student-$t$
intervals across checkpoints.
At the changed query, $h/\delta$ is 2 on the uniform grid and 1.2 on the stronger grid.
Protocols: Appendix~\ref{app:review-comparisons}.}
\label{tab:narrative-primary}
\begin{tabular}{lllc}
\toprule
State grid & Checkpoints & FID gain [95\% interval] & FID/KID wins \\
\midrule
Uniform & Prospective five & 1.906 $[1.772,2.040]$ & 5/5; 5/5 \\
Uniform & All seven & 2.023 $[1.820,2.226]$ & 7/7; 7/7 \\
Frozen stronger grid & Same five & 1.015 $[0.922,1.108]$ & 5/5; 5/5 \\
Frozen stronger grid & All seven & 1.041 $[0.960,1.122]$ & 7/7; 7/7 \\
\bottomrule
\end{tabular}
\end{table}

\subsection{Fixed-rule transfer and inference cost}

Without target-dataset retuning, the frozen $\rho=(1,1,1.2,1)$ rule improves FID on 9/9 historical CIFAR checkpoints across three recipes, all three evaluated public ImageNet checkpoints, and all four evaluated AFHQ checkpoints. Appendix Tables~\ref{tab:imagenet-absolute} and~\ref{tab:cross-data} report the original fixed-rule comparisons and distinguish them from dataset-tuned settings.

A fresh 50k confirmation gives small FID gains over Local across all three ImageNet checkpoints ($\SuppFixedXLTwoHGain$--$\SuppFixedBTwoHGain$), while IS decreases (Appendix~\ref{app:supplements-20260921}). These tests assess fixed-rule transfer; Section~\ref{sec:practical-calibration} compares target-calibrated temporal controls.

\paragraph{Inference cost.}
Interleaved, synchronized timing gives Local/Horizon batch latencies of $\CifarLatencyStandard/\CifarLatencyOurs$ ms on CIFAR and $\ImageNetLatencyStandard/\ImageNetLatencyOurs$ ms on ImageNet, with unchanged peak memory (Appendix Table~\ref{tab:latency-full}). Both use four velocity queries: four backbone calls on ImageNet at external CFG 1, or eight with two-pass CFG. Calibration and diagnostic costs are separate (Section~\ref{sec:practical-calibration} and Appendix~\ref{app:reproduction}).

\begin{figure}[htbp]
\centering
\includegraphics[width=\linewidth]{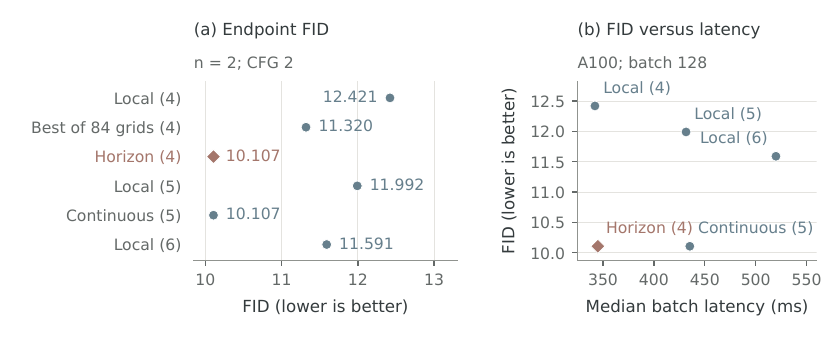}
\caption{The original two-checkpoint CFG-2 comparison illustrates comparable
Local/Horizon inference costs, not a broad Pareto frontier. (a) FID values match
Table~\ref{tab:cifar-main}B; query counts appear in parentheses. (b) The five
methods with measured median latency use Table~\ref{tab:latency-full}
(batch 128); the selected grid has no measured timing point. Lower FID is
better. Calibrated Start/Horizon at CFG 1.5 are a separate protocol
(Table~\ref{tab:narrative-calibration}).}
\label{fig:quality-latency}
\end{figure}

Figure~\ref{fig:quality-latency} relates the original two-checkpoint FID comparison to measured inference latency.

\section{Position Reversal and an Incoming-State Control}
\label{sec:horizon-response}

We now vary query position and use an incoming-state control to test whether time alone determines the response.

\subsection{The position reversal}

Figure~\ref{fig:time-horizon} isolates one query position at a time at CFG 2, with $h\geq\delta=.25$ and 10k paired samples per configuration. At $t=0$, extending $h$ from $.25$ towards $1$ worsens mean FID by up to $9.732$, with 0/7 wins at every tested non-local horizon. At $t=.25$, mean FID also worsens as the horizon increases. At $t=.5$, every tested non-local horizon improves FID on 7/7 checkpoints; $h=.50$ gives a mean improvement of $2.010$. The position scan is distinct from the 50k frozen-rule endpoint in Section~\ref{sec:efficacy}.

The source and midpoint queries have opposite responses at the same ratio, $\rho=2$. Thus a global horizon multiplier does not summarize the effect; where the longer horizon is applied matters.

Longer horizons therefore help selectively along this guided trajectory. Appendix~\ref{app:horizon-scope} places this scan alongside the existing short-horizon candidates and mixed CFG-1 midpoint results; an unguided position reversal remains unestablished. We next test whether earlier sampling choices change the midpoint benefit.

\subsection{Earlier guidance as a supporting control}
\label{sec:prefix-mechanism}

The position scan changes both query time and the states encountered at that time. We therefore keep the midpoint query time fixed and alter only earlier guidance, changing the distribution of states entering the midpoint. Within each prefix condition, Local and Horizon share the same entering states.

We use guidance $(p,p,1.5,1.5)$ with $p\in\{1,1.5\}$ and compare midpoint horizons $h_2\in\{.25,.50\}$. Only the first two guidance values vary. Current and subsequent guidance, the state-update grid, query start and model are held fixed. Define the paired horizon benefit and its prefix contrast as
\begin{equation}
G_p=\mathrm{FID}_{p,.25}-\mathrm{FID}_{p,.50},\qquad D=G_{1.5}-G_1.
\label{eq:prefix-contrast}
\end{equation}
The two-checkpoint pilot motivates $D>0$. We freeze this hypothesis before inspecting factorial outcomes on five additional checkpoints previously used in other horizon analyses. The contrast is positive on all five. Mean horizon benefit rises from $\PrefixUnguidedGain$ to $\PrefixGuidedGain$; mean $D$ is $\PrefixFiveGain$ with a descriptive 95\% Student-$t$ interval across checkpoints $[\PrefixFiveLow,\PrefixFiveHigh]$. KID contrasts have the same sign (Appendix~\ref{app:prefix-mechanism}).

A cross-recipe follow-up also finds positive FID/KID interactions on six historical checkpoints. Across its twelve Horizon-versus-Local comparisons, precision and class-label agreement decrease while recall increases (Appendix~\ref{app:prefix-mechanism}).

\paragraph{What the intervention identifies.}
The benefit of the midpoint intervention depends on the distribution of states reaching that query, beyond its time coordinate alone. Because earlier guidance changes those states and FID is nonlinear, the interaction does not distinguish CFG construction bias, horizon-dependent prediction error and other state-dependent effects. Its direction is also inconsistent across the ImageNet repeats (Appendix~\ref{app:public-prefix-followup}).

\section{Calibration Against Alternative Temporal Controls}
\label{sec:practical-calibration}

The preceding results establish a reproducible horizon effect. We now ask when it is worth calibrating this coordinate once query-start and state-grid controls are also available. Historical CIFAR, public ImageNet and new-seed tests compare these choices under stated candidate and sample budgets, matching new calibration cost within each comparison.

We compare three coordinates: the state grid $T$, query start $\tau$ and prediction horizon $h$. Start varies $\tau_k$ with $h_k=\delta_k$; Horizon varies $h_k$ with $\tau_k=t_k$. Both keep $T$ fixed, while Joint varies both query coordinates. Start follows prior temporal-conditioning calibration~\citep{li2024timeshift,xia2024timetuner}.

\subsection{Historical CIFAR: Start leads on FID, but evaluator rankings differ}
\label{sec:time-controls}
\label{sec:cfg-controls}

Historical CIFAR calibration favors Start. At CFG 1.5, Horizon reduces FID relative to Local by $\EndpointHorizonGain$, but Start reduces it by a further $\EndpointStartGain$; FID/KID improve on 7/7 checkpoints in both comparisons (Table~\ref{tab:narrative-calibration}A). Each family receives 25 candidates with 10k paired development samples per candidate, and its winner is frozen before fresh 50k evaluation on seven reused checkpoints. Candidate ranges inherit earlier discovery, whose costs are not matched.

% Editorial split of unchanged panels B/C in sections/reviewed_controls_20260924.tex.
% Sources and exact displayed values: evidence/narrative_20260926/table_provenance.json.
\begin{table}[t]
\centering
\small
\setlength{\tabcolsep}{4pt}
\caption{\textbf{Start leads on FID/KID, while Horizon yields lower DINO distance.}
Means over seven reused checkpoints; four queries, CFG 1.5.
Generated/real counts are 50k/50k in A and separate 10k/10k sets in B;
compare within panels.
B uses identical image bytes across evaluators.
P/R denote DINO precision/recall; dashes denote unmeasured metrics.
Details: Appendices~\ref{app:temporal-50k} and~\ref{app:samepixel-20260923}.}
\label{tab:narrative-calibration}
\begin{tabular}{lrrrrr}
\toprule
Rule & FID $\downarrow$ & KID $\times10^3\downarrow$ & DINO $\downarrow$ & DINO P & DINO R \\
\midrule
\multicolumn{6}{l}{\textit{A: Calibrated rules, CFG 1.5; 50k generated / 50k real}} \\
Local & 10.272 & 5.782 & -- & -- & -- \\
Calibrated Horizon & 8.878 & 4.831 & -- & -- & -- \\
Calibrated Start & \textbf{8.311} & \textbf{3.538} & -- & -- & -- \\
\midrule
\multicolumn{6}{l}{\textit{B: Identical saved pixels, CFG 1.5; 10k generated / 10k real}} \\
Local & 14.212 & 5.811 & 150.730 & 0.7171 & 0.3617 \\
Calibrated Horizon & 12.914 & 4.913 & \textbf{147.598} & 0.6967 & 0.3921 \\
Calibrated Start & \textbf{12.388} & \textbf{3.514} & 165.420 & 0.6821 & 0.3649 \\
\bottomrule
\end{tabular}
\end{table}

The ranking changes under DINOv2~\citep{oquab2023dinov2}. Table~\ref{tab:narrative-calibration}B evaluates the fixed calibrated choices on 21 saved 10k image sets, with identical rounded uint8 inputs to Inception and DINO. Horizon lowers FID/KID and DINO distance relative to Local on all seven checkpoints, while lowering precision and raising recall. Start has lower FID/KID than Horizon, but Horizon has lower DINO distance and higher precision/recall than Start on all seven. These measurements concern complete evaluators, including their different preprocessing, and come from separate 10k samples rather than panel A's 50k endpoints. Appendix~\ref{app:samepixel-20260923} reports checkpoint contrasts and quantization sensitivity.

The evaluator disagreement prevents treating calibration as a metric-independent ranking. Horizon has lower DINO distance than Start despite worse FID/KID, and its improvement over Local coincides with higher recall and lower precision. We therefore report the evaluator and precision--recall trade-off explicitly rather than treating a single distance as a complete ordering of generation quality.

\subsection{Public ImageNet: no added benefit from the tested Horizon allocation}

Target-calibrated ImageNet comparisons at external CFG 1 favor Start in both tested one-dimensional grids. A nine-point joint grid also selects the existing Start setting on both B/2 and XL/2 (Appendices~\ref{app:supplements-20260921} and~\ref{app:joint2d-20260922}). We next test Horizon candidates when both grid and query start can vary.

On public MeanFlow-B/2 and XL/2, Grid+Start+Horizon (GSH) has higher mean confirmation FID than Grid+Start (GS), failing the prespecified benefit criterion (Table~\ref{tab:public-grid-increment}). Each allocation receives 12 frozen candidates and 40k nominal development images per model; two fresh paired 10k repeats confirm the selected settings. Restoring the selected B/2 setting to its local horizon worsens FID by $0.0245$, below the $0.05$ threshold; XL/2 selects the Start anchor. GSH is not a strict superset of GS: this tests finite candidate allocations at equal new cost, not continuous-space optima (Appendix~\ref{app:public-grid-increment}).

% Generated from sealed confirmation reports by repro/build_public_grid_evidence_20260926.py.
\begin{table}[H]
\centering\small
\setlength{\tabcolsep}{8pt}
\begin{tabular}{lrrr}
\toprule
Model & Grid+Start & Grid+Start+Horizon & Restored $h=\delta$\\
\midrule
B/2 & 7.2227 & 7.2498 & 7.2743\\
XL/2 & 4.7969 & 4.8327 & 4.8327\\
\bottomrule
\end{tabular}
\caption{\textbf{The tested Horizon allocation adds no FID benefit.}
Mean FID ($\downarrow$) over two paired 10k confirmations, using four queries
and external CFG 1 with learned guidance retained.
GS and GSH have equal new budgets but nonnested candidate sets.
The last column restores $h=\delta$ in the selected GSH setting,
keeping its grid and query starts fixed.
Protocol: Appendix~\ref{app:public-grid-increment}.}
\label{tab:public-grid-increment}
\end{table}

\subsection{New training seeds: preferences change with query budget}
\label{sec:blind-nfe}

On two new CIFAR training seeds, Mixed selects Horizon at four queries, Start at eight and Local at two (Table~\ref{tab:blind-nfe-20260925}). At CFG 1.5, only the midpoint query changes. Mixed includes Local, two Start and two Horizon candidates, frozen using historical models before new-model rollouts. Mixed, Start-only and Horizon-only each select by FID from five candidates with 5k target images per candidate; two fresh paired 10k repeats confirm the frozen choices. Appendix~\ref{app:blind-nfe-20260925} gives the candidate allocation and full accounting.

\begin{table}[htbp]
\centering
\small
\setlength{\tabcolsep}{4pt}
\begin{tabular}{lrrrrrl}
\toprule
Seed & NFE & Local & Fixed H & Cal. H & Cal. Start & Mixed (choice) \\
\midrule
700081 & 2 & 15.122 & n/a & 15.122 & 15.122 & 15.122 (Local) \\
700081 & 4 & 14.123 & 13.670 & 12.632 & 12.844 & 12.632 (H) \\
700081 & 8 & 13.061 & 13.009 & 12.771 & 12.680 & 12.680 (Start) \\
700087 & 2 & 15.075 & n/a & 15.075 & 15.075 & 15.075 (Local) \\
700087 & 4 & 14.326 & 13.848 & 12.648 & 12.912 & 12.648 (H) \\
700087 & 8 & 13.288 & 13.234 & 12.997 & 12.989 & 12.989 (Start) \\
\bottomrule
\end{tabular}
\caption{\textbf{On new training seeds, the FID-selected control changes with NFE.}
Mean FID ($\downarrow$) over two paired 10k confirmations at CFG 1.5,
against 10k real images.
Fixed H uses midpoint $h/\delta=1.2$; this extension is infeasible at NFE 2 (n/a).
Cal.\ H/Start select within the named family plus Local;
Mixed includes both families, with its choice in parentheses.
Candidate sets, budgets and other metrics: Appendix~\ref{app:blind-nfe-20260925}.}
\label{tab:blind-nfe-20260925}
\end{table}

At four queries, Horizon lowers FID in all four paired comparisons with calibrated Start and also lowers DINO distance. KID is higher than with Start; relative to Local, the drop in DINO precision exceeds the prespecified threshold, while recall rises. At eight queries, the selected Start worsens DINO distance relative to Local. At two queries, all calibrated strategies select Local; the fixed horizon extension is infeasible because the midpoint is the final update.

Horizon is therefore competitive at four queries on these two seeds, with a different preferred control at other budgets. The added value of learning the candidate allocation remains unresolved. Appendix~\ref{app:blind-nfe-20260925} reports the metric contrasts, fixed-rule outcomes and allocation analysis.

\section{Discussion}

\paragraph{Why the horizon matters.}
An exact interval-average predictor favors $h=\delta$ because the resulting update gives the intended flow displacement. Equation~\plaineqref{eq:error-decomposition} shows what changes for a learned model: varying $h$ changes both the mismatch between requested and physical intervals and the predictor's approximation error. The horizon therefore selects how the learned field is queried, in addition to the state grid's choice of integration steps. The position reversal and incoming-state control reveal structure in this combined response. They leave the relative roles of approximation error and CFG construction bias unresolved; exploratory directional and local-reference studies examine these possibilities in Appendices~\ref{app:mechanism-20260924} and~\ref{app:reference}.

The decomposition also clarifies why local prediction accuracy and endpoint performance need separate evaluation. Its two terms are vectors: changing their magnitudes alone does not determine the magnitude of their sum. An endpoint metric additionally evaluates the distribution after the remaining updates, which act on the states produced by the intervention. Agreement with a local reference and improvement of the final distribution therefore answer different questions. In our reference diagnostic, longer midpoint horizons move farther from both the native-local and instantaneous-Heun reference velocities even where endpoint FID improves (Appendix~\ref{app:reference}). These references are constructed from the learned model, not ground-truth transport trajectories. Their disagreement with FID cautions against using local agreement as a selection criterion on its own; it does not identify the error responsible for the endpoint gain.

\paragraph{Selecting the horizon.}
Once chosen, Horizon adds no velocity queries, but the tested selectors require target rollouts. The preferred temporal control changes across models, NFE values and evaluators, and the tested diagnostic predictor fails its prespecified success criterion (Appendix~\ref{app:prediction-20260922}). Practical selection therefore remains target-dependent.

More broadly, our experiments separate two questions: whether a model input changes endpoint behavior, and whether allocating calibration budget to that input improves on available alternatives. The position and incoming-state controls establish the former for prediction horizon; the matched-budget comparisons show that the latter is configuration dependent.

\paragraph{Scope of the finding.}
The position reversal and prospective replication are established within one guided CIFAR architecture and training recipe. The blind study adds two training seeds at NFE values already seen in development. Fixed-rule FID gains transfer to the listed public checkpoints, but native iMF interventions fail their benefit criterion (Appendix~\ref{app:imf-position-followup}). Together with the Start comparisons and metric trade-offs, this evidence supports configuration-specific horizon calibration. A transferable rule for choosing the horizon remains a separate research challenge.

\section{Related Work}

\paragraph{Few-step MeanFlow inference.}
MeanFlow predicts interval-average velocities~\citep{geng2025meanflow}; subsequent work studies training stability, guidance and gap curricula~\citep{geng2026improved,kim2026understanding}. MeanCache constructs interval averages through Jacobian--vector-product reuse~\citep{gao2026meancache}. Decoupled MeanFlow reassigns time conditioning across backbone blocks, studies conversion with and without fine-tuning, and uses restart/CTM samplers that can request an intermediate endpoint before a separate state transition~\citep{lee2026decoupled}. Our deterministic intervention retains the existing model interface, query start and physical grid.

\paragraph{Two-time denoisers and prediction horizon.}
\citet{lee2026language} introduce a two-time denoiser for language modeling. Flow Map Denoisers~\citep{zilberstein2026denoisers} use $D_{s,t}(x)=x+(1-s)v(x,s,t)$, with $v$ averaged over $[s,t]$, and vary $t$ to control the distortion--perception trade-off. This is an algebraic instance of Equation~\plaineqref{eq:decoupled-update} with $\delta=1-s$ and $h=t-s\leq\delta$. Our empirical focus is intermediate-step generation with $h>\delta$, especially the position reversal and calibration against Start. Existing short-horizon candidates are summarized in Appendix~\ref{app:horizon-scope}. The contribution is the controlled characterization of the response in pretrained MeanFlow; the algebraic separation itself is shared with earlier two-time models.

\paragraph{State-grid and query-time calibration.}
Align Your Steps, Optimal Stepsize Distillation, Optimize Your Sampling, and IntMeanFlow's O3S change the state/noise grid, with queries tied to state transitions~\citep{sabour2024align,pei2025optimal,zhang2026oys,wang2026intmeanflow}.  Shortcut Models similarly tie the predictor's step condition to the update span~\citep{frans2025shortcut}.  Time-Shift Sampler adjusts a denoiser's current-time condition~\citep{li2024timeshift}.  TimeTuner changes the queried time while retaining the update interval, and MTEO optimizes step- and layer-specific time embeddings while freezing the diffusion backbone~\citep{xia2024timetuner,cai2026three}.  Our intervention varies the prediction horizon of an interval-average predictor while fixing its query start and state grid.

\paragraph{Higher-order and adaptive numerical solvers.}
STORK designs stabilized Runge--Kutta methods for diffusion and flow-matching ODEs~\citep{tan2026stork}; ERK-Guid uses embedded Runge--Kutta local error and stiffness as a guidance signal~\citep{kong2026erk}.  These methods integrate an instantaneous field, whereas our method changes the queried interval-average prediction.  Appendix~\ref{app:discovery} points to query-count-matched diagnostics through MeanFlow's $h=0$ branch; these adapted ports are not native solver implementations.

\paragraph{Guidance along the trajectory.}
Limited-interval guidance studies the position dependence and accumulated effects of CFG~\citep{kynkaanniemi2024limited}. Our factorial diagnostic holds guidance fixed at and after the intervention, and measures how earlier CFG changes the benefit of a later horizon adjustment. The identified effect concerns the incoming-state distribution.

\section{Conclusion}

Prediction horizon is a separately adjustable inference coordinate in pretrained few-step MeanFlow. Its response is structured: longer horizons hurt near the source and help at the midpoint in our guided CIFAR setting, while earlier guidance changes the midpoint benefit. The useful midpoint effect replicates across training seeds and persists on a stronger frozen state grid. Comparisons with query-start and grid controls show that the preferred choice varies across the tested models, query budgets and evaluators. The evaluated selection procedures use target rollouts. Prediction horizon therefore expands the inference-time design space of few-step MeanFlow with fixed model weights and query count.
\label{page:main-text-end}
\section*{AI Use Statement}
AI coding and language assistants were used in manuscript drafting and editing.  Responsibility for the submitted work remains with the authors.

\section*{Reproducibility Statement}
The appendices specify the sampler, training recipes, paired evaluation, candidate budgets and confirmatory outcomes, including the comparisons that limit each claim. The source package contains frozen protocols, per-cell reports and portable table renderers; \path{repro/EVIDENCE_INDEX.md} locates the full historical records. Appendix~\ref{app:reproduction} distinguishes report checks from independent metric evaluation and lists the external assets needed for full reproduction. The scoped supplementary subtotal is now 3,812,000 endpoint images, including the 244,000-image public GS/GSH study; this is not the full project total. The package supports standalone manuscript compilation and the minimal sampler demonstration.

\bibliography{references_readthrough}
\bibliographystyle{iclr2027_conference}

\clearpage
\appendix
\label{page:appendix-start}
% Editorial consolidation only; no new experiment or numerical recomputation.
% Sources: inherited sections/appendix.tex (algorithm/protocol), supplement.tex
% (recipes/aggregation/derivation), reviewed_evidence_map_20260923.tex, and
% blind_nfe_appendix_20260925.tex. Complete originals remain in the source
% archive and archive/full_manuscript_20260925.pdf.
\section{Sampler, Training Recipes, and Evaluation Protocol}
\label{app:compact-methods}

The experiments distinguish the interval requested from the model, the duration
of each state update, and the computation used to select a schedule. Candidate
lists, reference sets and frozen choices are specified for each study; full
experimental records are indexed in the source package (Appendix~\ref{app:reproduction}).

\subsection{Sampler and model-interface contract}
\label{app:algorithm}
\label{app:minimal-sampler}

Let $T=(t_0,\ldots,t_K)$ be the physical state grid, $\tau_k$ the queried
start time, and $h_k$ the queried duration. The frozen model returns a velocity;
the physical update multiplies it by $\delta_k=t_{k+1}-t_k$, irrespective of
the requested duration. Horizon-only control sets $\tau_k=t_k$ and changes
$h_k$. Start-only control changes $\tau_k$ and retains $h_k=\delta_k$.
Neither changes $T$.

\begin{algorithm}[htbp]
\centering
\begin{minipage}{.92\linewidth}
\textbf{Input:} initial state $x_0$; increasing grid $T$ from 0 to 1;
query starts $\tau$ and durations $H$; frozen velocity callable $u_\theta$.

Validate $0\leq\tau_k\leq\tau_k+h_k\leq1$ for every query.

\textbf{For} $k=0,\ldots,K-1$:
\[
\delta_k\leftarrow t_{k+1}-t_k,\qquad
v_k\leftarrow u_\theta(x_k,\tau_k,h_k),\qquad
x_{k+1}\leftarrow x_k+\delta_k v_k.
\]
\textbf{Return} $x_K$. There are exactly $K$ velocity queries.
\end{minipage}
\caption{A fixed state grid with independently specified query intervals.
Local MeanFlow is recovered by $\tau_k=t_k$ and $h_k=\delta_k$.}
\label{alg:sampling}
\end{algorithm}

A duration-based interface receives $h_k$; an endpoint-based interface receives
$\tau_k+h_k$. A reverse-time checkpoint additionally needs its native time
and velocity convention, implemented by its adapter. The instantaneous limit
$h=0$ is legal only for an interface that supports it. Invalid schedules are
rejected, not clipped. In particular, the midpoint of a two-step uniform grid
is its final update, so extending its local horizon $.5$ is outside the query
domain. The new blind study therefore marks the fixed extension there as
inapplicable and returns Local.

Labels and guidance belong in the callable. For two-pass classifier-free
guidance, $v=v_{\rm null}+s(v_{\rm cond}-v_{\rm null})$, evaluated with identical
$(\tau,h)$ in both branches; $K$ velocity queries then cost $2K$ backbone
forwards. Public-model panels specify whether external or native guidance is
used and record actual calls. Once a duration schedule has been selected, Horizon control adds no model evaluation during deployment; calibration is a separate cost.

The standalone \path{repro/sampler.py} accepts a velocity callable, noise,
$T$, $\tau$, and $H$, returning the final state, trajectory and query count.
Run \texttt{python demo.py} from \path{repro/} for its CPU checks without
weights or data. The model used by the callable must be in evaluation mode; disabling
gradients alone does not disable dropout. Example schedules include Local,
the mild $h_2=.30$ rule, Start $(\tau_2,h_2)=(.575,.25)$, and the historical
calibrated Horizon $(.5,.4895833333)$ on the four-step uniform grid. These
examples implement the sampler, not automatic metric-based selection.

\subsection{Training recipes and checkpoint provenance}
\label{app:training-recipes}

The seven principal CIFAR checkpoints and the two later blind-study seeds use
the same class-conditional U-Net: 128 base channels, two residual blocks per
resolution, multipliers $(1,2,2,2)$, attention at resolution 16, four heads with
64 channels per head, dropout .1, and an explicit interval input. Training uses
the MeanFlow Jacobian--vector product (JVP) objective, 250 epochs, AdamW at $5\times10^{-4}$, random
horizontal flips, effective batch 256 and EMA decay .999. Final EMA weights
are evaluated. The registered two-rank implementation uses 128 samples per
rank and 195 optimizer steps per epoch, totaling 48,750 final optimizer steps.

\paragraph{Fixed map and online pairing.}
This is a project-specific training recipe. The fixed object is an offline
prototype map; sample pairs are recomputed from each local batch of 128.
Let $\phi$ be flattened $4\times4$ average-pooled RGB features and
$m(a)=\|a\|_2^2/\dim(a)$. The shared map (SHA-256 prefix
\texttt{7372efed6f05}) stores 16 source/target prototypes
$(a_{c\ell},b_{c\ell})$ per class, source-cell dual weights $\beta_{c\ell}$,
and a prototype permutation $\sigma_c$ obtained by endpoint-unary Hungarian
matching under a fixed seed-42 teacher. The dual weights fit source-cell
masses to the mapped target-cluster frequencies. Define
\[
\ell_i=\arg\min_\ell\{\|\phi(z_i)-a_{c_i\ell}\|_2^2-\beta_{c_i\ell}\},
\quad B_i=b_{c_i,\sigma_{c_i}(\ell_i)},\quad A_i=a_{c_i\ell_i}.
\]
For Gaussian noise $z_i$ and candidate image $x_j$ of the same class, the
map contributes the fixed guidance cost
\[
G_{ij}=R(\phi(x_j)-\phi(z_i),B_i-A_i)+.5R(\phi(x_j),B_i),
\qquad R(a,b)=\frac{m(a-b)}{m(a)+m(b)+10^{-12}}.
\]
The current EMA supplies instantaneous endpoint velocities $v_i^0,v_j^1$;
with $d_{ij}=x_j-z_i$, the online cost is
$C_{ij}=\{m(v_i^0-d_{ij})+m(v_j^1-d_{ij})\}/2$, normalized over the full local batch before class masking as
$\widetilde C=C/\max(\max_{ij}C_{ij},10^{-12})$.
Let $\pi_0$ minimize this unary cost over class-preserving permutations and
$U_c(\pi)=\sum_{i:c_i=c}\widetilde C_{i,\pi(i)}$, $U=\sum_c U_c$.
The candidate graph retains $\pi_0$ and up to eight lowest-cost same-class
edges per source under $\widetilde C+G$. The solver targets
\[
\sum_iG_{i,\pi(i)}+950\frac{U(\pi)}{\max(U(\pi_0),10^{-12})},
\qquad U_c(\pi)\leq1.02\,U_c(\pi_0)\quad\text{for every class }c.
\]
It selects the best visited feasible assignment from deterministic Hungarian
solves with a classwise multiplier search (when needed, up to 64 bracket
doublings followed by 32 bisections); it does not certify
the constrained global optimum. Thus .02 is a classwise unary-cost allowance
and 950 is a scalarization weight, not a matching constraint penalty.
The online quadratic relation weight is zero. Changing rank count and local
batch size would change these matching pools.

\paragraph{Training intervals and loss.}
Independently draw $a\sim\mathcal N(.4,1)$ and $b\sim\mathcal N(-.4,1)$,
set $t=\min(\operatorname{sigmoid}(a),\operatorname{sigmoid}(b))$ and
$r=\max(\operatorname{sigmoid}(a),\operatorname{sigmoid}(b))$, then set $t=r$
with probability .75. Thus the instantaneous branch has probability .75
and positive intervals probability .25. For each paired $(z,x)$, use
$x_t=(1-t)z+tx$, $v=x-z$,
and $\widetilde u_\theta(x_t,t,r)=u_\theta(x_t,t,r-t)$. The JVP target is
$\operatorname{stopgrad}[v+(r-t)(v\cdot\nabla_x+\partial_t)\widetilde u_\theta]$,
where $r$ is held fixed. If $S$ is the squared $\ell_2$ prediction error,
the loss averages $S/(\operatorname{stopgrad}S+10^{-3})^{.75}$ over the batch;
class labels are independently dropped with probability .1.
The frozen source locations and verified hashes are recorded in
\path{evidence/revision_20260925/training_definition_sources.json}.

\begin{table}[htbp]
\centering\footnotesize
\setlength{\tabcolsep}{4pt}
\begin{tabular}{p{.24\linewidth}p{.35\linewidth}p{.31\linewidth}}
\toprule
Cohort & Checkpoints & Role and replication unit \\
\midrule
Original CIFAR & 300007, 300017 & Two training units; three paired 50k repeats each \\
Prospective extension & 700001, 700027, 700057, 700067, 700079 & Five new training units; one paired 50k repeat each \\
Later blind study & 700081, 700087 & Two separate new units; NFE 2/4/8, two 10k confirmations per selected setting \\
Historical CIFAR transfer & Seeds 42, 123, 2027 in each of three pairing recipes & Reused historical models; separate transfer evidence \\
Public ImageNet & MeanFlow-B/2, MeanFlow-XL/2, AlphaFlow-B/2 & Three public checkpoints, not common-architecture seed replication \\
AFHQv2 & Two corrected and two QMAC-trained checkpoints & Four held-out checkpoints in two training families \\
\bottomrule
\end{tabular}
\caption{Evidence cohorts are kept separate. Additional metrics or generation
repeats do not increase the number of independently trained models.}
\label{tab:compact-evidence-units}
\label{tab:reviewed-evidence-map}
\label{tab:evidence-design}
\end{table}

Table~\ref{tab:compact-evidence-units} summarizes the evidence cohorts. The historical CIFAR families share the U-Net and training budget but use
online model-aligned matching (called MAC-paired below; artifact name \texttt{official}), deterministic
class-matched unary assignment (\texttt{corrected}), or a fixed semidiscrete
map. Their earlier availability prevents treating them as blind transfer.
For AFHQ, the labels corrected and QMAC-trained denote unary matching and
relation-aware matching, respectively. AFHQ uses a latent DiT-B/2 on $4\times32\times32$ latents, 900 training epochs,
effective batch 256, AdamW at $5\times10^{-4}$ and final EMA .999. The
\path{stabilityai/sd-vae-ft-mse} VAE decodes to $256\times256$ images. The three public ImageNet
checkpoints come from the AlphaFlow release; source locations, complete
checkpoint digests and adapters are preserved in the archived protocols.
The later blind study records its same-recipe checkpoint resumption separately
and does not claim an uninterrupted, bitwise-identical RNG trajectory.

\subsection{Paired evaluation, independent units and audit scope}
\label{app:protocol}
\label{app:editorial-support-20260923}

The primary CIFAR rule and position scan use CFG 2; temporal calibration and
the new blind study use CFG 1.5. The primary endpoints use 50k generated images
and the complete 50k CIFAR training reference. Position scans use paired 10k
cells. Historical calibration has separate 10k and 50k endpoints; the blind-NFE
study uses 2k old-model development, 5k target calibration and paired 10k
confirmation, with a fixed 10k real reference. Absolute FID values across
these protocols are not directly comparable.

Paired arms share the initial noise, labels, checkpoint, sample count and
evaluation environment. Within a comparison only the specified query, grid,
guidance or velocity intervention changes. For a training seed $s$ with
$n_s$ generation repeats, aggregation is
\[
\bar M_{s,a}=n_s^{-1}\sum_g M_{s,g,a},\qquad
\bar M_a=|\mathcal S|^{-1}\sum_{s\in\mathcal S}\bar M_{s,a}.
\]
The seven-checkpoint estimates therefore give each training seed equal weight,
despite differing repeat counts. Reported gains are computed from unrounded scores. Their Student-$t$ intervals describe
checkpoint variation; pointwise surface intervals are not simultaneous bands.
The two new seeds remain a distinct cohort. Their prespecified distance gates
are descriptive criteria, not training-population significance tests.

FID/KID use the frozen Inception evaluator and matching real statistics within
each comparison. CIFAR KID uses 100 subsets of 1,000 samples; the AFHQ tuned
confirmation uses 20 subsets of 500 (Appendix~\ref{app:supplements-20260921}).
The original CIFAR environment records PyTorch 2.3.1, TorchMetrics 1.8.2 and
torch-fidelity 0.4.0. Historical Inception truncation and subsequent rounded
uint8 reevaluation remain distinguished. The same-pixel audit and blind
confirmation feed identical rounded image bytes to Inception and DINOv2;
feature resizing and normalization still differ. DINO precision/recall uses
the named feature representation and $k=3$. VGG and DINO P/R are not
interchangeable; FID, KID, DINO distance and IS do not imply uniform perceptual
or semantic improvement.

The direct four-query schedule was selected on historical seed 42 and frozen
for its three paired 50k comparisons on seeds 300007/300017. Those two weights
also occur in the preceding sparse-correction studies and later development
panels; freezing this comparison does not make them unseen throughout the
discovery process. The additional five-model protocol was frozen before
training. Fixed-grid and later temporal panels state their reuse explicitly.
In the blind-NFE study (Appendix~\ref{app:blind-nfe-20260925}), candidate allocation uses only old-model development, while each target choice is sealed before confirmation. Equal target calibration budgets exclude old development and historical discovery. Image reuse adds neither training units nor generated
images.

\subsection{Reproduction scope}
\label{app:reproduction}
\label{app:archive}
The source package compiles independently and includes the CPU sampler demo,
portable numerical renderers, frozen protocols and per-cell reports. The
historical report in \path{archive/full_manuscript_20260925.pdf} preserves full
candidate tables and individual repeats; its appendix and table numbers belong
to that edition. \path{repro/EVIDENCE_INDEX.md} provides the detailed record map,
accounting and build instructions.

Full training, generation or metric reproduction additionally requires the
weights, datasets, pairing map, image/feature arrays and dependencies listed in
\path{repro/ASSET_ACCESS.md}. Numerical renderers reconstruct tables from stored
reports; they do not independently evaluate images. Audit coverage is specific
to each study: the directional panels include cached-feature FID/DINO
recomputation, while the blind-study completion audit checks report arithmetic,
pairing, selection chronology and file bindings without re-extracting features.

The named supplementary panels total 3,812,000 endpoint images: 2,350,000 from seven earlier panels, 480,000 from directional studies, 738,000 from the blind-NFE study, and 244,000 from the public GS/GSH study (Appendix~\ref{app:public-grid-increment}). This is a scoped subtotal, not the project
total; training, diagnostic probes and feature extraction are accounted for
separately. Reused images and sampling repeats add no independent training units.

\subsection{Average-velocity expansion and the update error}
\label{app:derivations}

Along a smooth trajectory, with $D/Dt$ the material derivative,
\begin{equation}
\bar v_h(t)=v_t+\frac h2\frac{Dv_t}{Dt}
             +\frac{h^2}{6}\frac{D^2v_t}{Dt^2}+O(h^3).
\label{eq:appendix-average-expansion}
\end{equation}
Writing $u_\theta(x_t,t,h)=\bar v_h(t)+\epsilon_\theta(t,h)$, the displacement
error when updating by $\delta$ is
\begin{equation}
e_{\delta,h}(t)=\delta\left[
\frac{h-\delta}{2}\frac{Dv_t}{Dt}
+\frac{h^2-\delta^2}{6}\frac{D^2v_t}{Dt^2}
+\epsilon_\theta(t,h)+O(h^3+\delta^3)\right].
\label{eq:appendix-displacement-error}
\end{equation}
Thus changing $h$ changes both trajectory mismatch and the finite predictor's
error. Neither term is identified by the endpoint FID comparisons, and the
expansion supplies no guarantee or rule for predicting the useful horizon.
Changing $T$ changes the update times and durations. Changing $H$ at fixed
$T$ changes the requested prediction intervals while retaining those update
times and durations; the resulting velocities and subsequent states can still
differ. The standard coupling $H=\Delta T$ restricts the allowed $(T,H)$ schedules, regardless of whether $T$ is uniform.

\paragraph{External CFG and the reference flow.}
In Equation~\plaineqref{eq:error-decomposition}, $u_\theta$ denotes the effective
predictor, including external guidance. Its error can contain guidance
construction bias even if each branch is exact. For example, let the
conditional and null instantaneous fields be $ax$ and $bx$. Their exact
interval averages are $(e^{ah}-1)x/h$ and $(e^{bh}-1)x/h$. Mixing these with
weights $w,1-w$ differs from the average of the guided field
$[wa+(1-w)b]x$ by
\[
\frac h2 w(1-w)(a-b)^2x+O(h^2).
\]
Thus learning the guided average directly, as in the MeanFlow guidance
construction~\citep{geng2025meanflow}, and mixing branch averages are distinct.
The decomposition remains valid with this bias included in $\epsilon_\theta$;
this analytic example does not identify the source of our empirical gains.
\FloatBarrier

% Editorial consolidation only. Numeric rows are copied from inherited
% reviewed_primary_rows_20260923.tex and generated_appendix_tables.tex;
% aggregate macros and latency values retain their original generated sources.
% Qualitative images are paired historical evaluator previews compressed for display.
% Full repeated-cell tables remain in archive/full_manuscript_20260925.pdf.
\section{Primary Replication, Fixed-Grid Control and Position Effects}
\label{app:review-comparisons}

The core results use CFG 2 and are distinct from the historical CFG-1.5
calibration and later blind-NFE study. Training checkpoints are the independent
units for the replication, while paired generation repeats measure variability
within a checkpoint.

\subsection{Seven-checkpoint replication and the stronger grid}

On the uniform four-step grid, the primary rule changes only $h_2$ from $.25$
to $.50$. It improves FID and KID on all seven checkpoints. Repeats are first
averaged within training seed: three for each original seed and one for each
prospectively added seed. The five-model subgroup alone retains the effect
(Table~\ref{tab:cifar-main}A); it is not inferred from the combined mean.
The seven-checkpoint uniform-grid Local/Horizon means are FID $12.241/10.218$
and KID $\times10^3$ $5.682/4.226$.

\begin{table}[htbp]
\centering\small
\caption{\textbf{Primary efficacy and fixed-grid control at CFG 2.}
A separates the prospective subgroup from the combined cohort. Intervals are
descriptive 95\% Student-$t$ intervals across checkpoints. B is the original two-checkpoint
head-to-head comparison, with three paired 50k repeats per checkpoint.
``Best of 84'' denotes a finite-grid winner, not a family optimum.}
\label{tab:cifar-main}
\textit{A: Checkpoint aggregates after fixing the state grid}

\resizebox{\linewidth}{!}{\begin{tabular}{llrrlc}
\toprule
State grid & Checkpoints & Local FID & Horizon FID & Gain [95\% interval] & FID/KID wins \\
\midrule
Uniform & Prospective five & 12.168 & 10.262 & 1.906 $[1.772,2.040]$ & 5/5; 5/5 \\
Uniform & All seven & 12.241 & 10.218 & 2.023 $[1.820,2.226]$ & 7/7; 7/7 \\
Best of 84 & Same five & 11.197 & 10.182 & 1.015 $[0.922,1.108]$ & 5/5; 5/5 \\
Best of 84 & All seven & 11.236 & 10.195 & 1.041 $[0.960,1.122]$ & 7/7; 7/7 \\
\bottomrule
\end{tabular}
}

\vspace{4pt}
\textit{B: Original two-checkpoint head-to-head means}

\begin{tabular}{lclrr}
\toprule
Method & Queries & State grid & FID $\downarrow$ & KID $\downarrow$ \\
\midrule
Local MeanFlow & 4 & uniform & \CifarStandardFID & \CifarStandardKID \\
Best of 84 grids & 4 & selected & \CifarTimestepFID & \CifarTimestepKID \\
Horizon & 4 & uniform & \CifarOursFID & \CifarOursKID \\
Local MeanFlow & 5 & uniform & \CifarNFEFiveFID & \CifarNFEFiveKID \\
Continuous lookahead & 5 & uniform & \CifarContinuousFID & \CifarContinuousKID \\
Local MeanFlow & 6 & uniform & \CifarNFESixFID & \CifarNFESixKID \\
\bottomrule
\end{tabular}
\end{table}

For the stronger-grid control, development evaluates the 84 triples of
internal knots from $\{.1,.2,\ldots,.9\}$. Its winner
$T^\star=(0,.3,.6,.9,1)$ is independently rechecked and frozen before
follow-up. The state-update durations remain $(.3,.3,.3,.1)$; Horizon changes only
the third prediction horizon from $.30$ to $.36$, using the frozen ratio 1.2. No joint grid/horizon
search is performed in this follow-up. Across the seven checkpoints, the mean gains are
$\JointTHGain$ FID and $\JointTHKIDGain$ KID; the five-model subgroup retains
$\JointTHNewGain$ FID. Table~\ref{tab:seven-cifar-checkpoints} reports the paired effects on all seven checkpoints.

\begin{table}[htbp]
\centering\footnotesize
\setlength{\tabcolsep}{4pt}
\begin{tabular}{rrlrrrr}
\toprule
Train seed & \shortstack{Uniform\\repeats} & Cohort & \multicolumn{2}{c}{Uniform-grid gain}
& \multicolumn{2}{c}{Frozen-grid gain} \\
& & & FID & KID $\times10^3$ & FID & KID $\times10^3$ \\
\midrule
300007 & 3 & original & 2.3628 & 1.755 & 1.1835 & .794 \\
300017 & 3 & original & 2.2657 & 1.629 & 1.0269 & .674 \\
700001 & 1 & prospective & 1.8780 & 1.374 & .9956 & .665 \\
700027 & 1 & prospective & 1.7943 & 1.317 & .9203 & .574 \\
700057 & 1 & prospective & 1.8535 & 1.354 & .9859 & .657 \\
700067 & 1 & prospective & 1.9242 & 1.228 & 1.1176 & .721 \\
700079 & 1 & prospective & 2.0798 & 1.536 & 1.0564 & .687 \\
\midrule
Mean & & seven models & 2.0226 & 1.456 & 1.0409 & .682 \\
\bottomrule
\end{tabular}
\caption{Paired Local-minus-Horizon effects on the same seven training
checkpoints. Uniform and frozen-grid protocols are separate comparisons. Frozen-grid effects use one 50k repeat per checkpoint.
Per-arm absolute scores and every generation-repeat row remain in the archived
full tables; their rounded differences are not used to recompute aggregates.}
\label{tab:seven-cifar-checkpoints}
\label{tab:joint-th-replication}
\end{table}

\subsection{Position dependence and the discovery control}
\label{app:discovery}

The direct four-query rule was selected by minimum 10k FID on historical
seed 42 from eight binary schedules: each of the first three queries used
either its local duration or the remaining interval. The selected schedule
changes only the midpoint to $h=.50$. Its settings and three generation seeds
were fixed for the subsequent 50k comparison on 300007/300017; these two
weights also appeared in earlier sparse-correction studies. The later five
training seeds were separately registered before training.

The mild ratio $\rho=1.2$ comes from the smallest non-local midpoint ratio in
the pre-existing CIFAR sweep. It was frozen as a conservative transfer rule,
without target-dataset search, rather than chosen as a per-dataset optimum.

The position scan changes one horizon at a time while preserving the uniform
four-update grid. Each cell uses paired 10k images on each of the seven
checkpoints at CFG 2. Table~\ref{tab:time-horizon-seven} retains the full
checkpoint-mean surface from the main figure, including harmful early
interventions. The pointwise intervals in Figure~\ref{fig:time-horizon} are not simultaneous
confidence bands.

\begin{table}[htbp]
\centering\scriptsize
\setlength{\tabcolsep}{3pt}
\begin{tabular}{cp{.24\linewidth}p{.37\linewidth}p{.20\linewidth}}
\toprule
State time & Queried horizons & Mean FID gains & Positive checkpoints \\
\midrule
0 & .25,.375,.50,.625,.75,.875,1
& .000,-.798,-2.220,-3.855,-5.639,-7.531,-9.732
& 0/7,0/7,0/7,0/7,0/7,0/7,0/7 \\
.25 & .25,.375,.50,.625,.75
& .000,-.031,-.092,-.280,-.837
& 0/7,2/7,1/7,1/7,0/7 \\
.50 & .25,.30,.35,.40,.45,.50
& .000,+.489,+.970,+1.414,+1.774,+2.010
& 0/7,7/7,7/7,7/7,7/7,7/7 \\
\bottomrule
\end{tabular}
\caption{Full position-scan means. Positive gain favors the intervention.
The local $h=.25$ entries are exact zero contrasts, not failed replications.}
\label{tab:time-horizon-seven}
\end{table}

The earlier continuous family queries both a local and longer velocity and
uses $v_s+\alpha(v_l-v_s)$. Its selected $\alpha=1$ reduces algebraically to
$v_l$, allowing the redundant local query to be removed. Thus Table~\ref{tab:cifar-main}B
compares a five-query discovery arm with the four-query rule; it is not an
extra inference operation required by Horizon. Independently materialized
evaluation paths can differ in floating-point order despite this algebraic
identity. The archived full report's Appendix C.2, ``Public schedule and solver
ports,'' gives the query-matched instantaneous-field ports and their results
(Table 28 in that archived edition; source label \texttt{tab:public-ports}). These
$h=0$ interface diagnostics are distinct from the native-interval best-of-84
grid control; their poor results do not establish a general solver ranking.

A separate CFG-2 VGG-feature audit on the original two checkpoints shows the
quality--coverage boundary: Local precision/recall is $.7509/.3547$, versus
$.6691/.4830$ for Horizon. It uses paired 10k samples and does not measure the
50k endpoints. Lower FID here therefore does not establish uniformly improved
perceptual quality.

\subsection{Inference latency}

Timing uses an A100-PCIE-40GB, fixed batch and precision, warm-up,
interleaved modes and CUDA synchronization around each timed region. It is
separate from generation logs that include metrics or changing GPU occupancy.
Table~\ref{tab:latency-full} reports median batch time; the source audit retains
the p05/p95 values. Peak allocated memory is unchanged between Local and
Horizon in both datasets. The small negative ImageNet difference is consistent
with measurement noise. Figure~\ref{fig:quality-latency} relates these timings to the original two-checkpoint FID comparison.

\begin{table}[htbp]
\centering\small
\begin{tabular}{lrrr}
\toprule
Sampler & Velocity queries & CIFAR ms (batch 128) & ImageNet ms (batch 16) \\
\midrule
Local & 4 & 342.09 & 212.71 \\
Horizon & 4 & 344.93 & 212.54 \\
Local & 5 & 431.59 & 226.56 \\
Continuous lookahead & 5 & 435.29 & 225.35 \\
Local & 6 & 519.91 & 238.95 \\
\bottomrule
\end{tabular}
\caption{Interleaved median inference latency, copied from the frozen timing
audit. Horizon changes the four-query median by +.83\% on CIFAR and -.08\%
on ImageNet. Query counts are velocity evaluations; CFG call accounting is
specified by the corresponding model protocol.}
\label{tab:latency-full}
\end{table}

\subsection{Paired qualitative previews}
\label{app:qualitative}

\begin{figure}[htbp]
\centering
\begin{tabular}{ccc}
\textbf{CIFAR-10} & \textbf{ImageNet-256} & \textbf{AFHQv2-256} \\
\includegraphics[width=.30\linewidth]{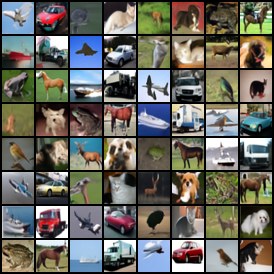} &
\includegraphics[width=.30\linewidth]{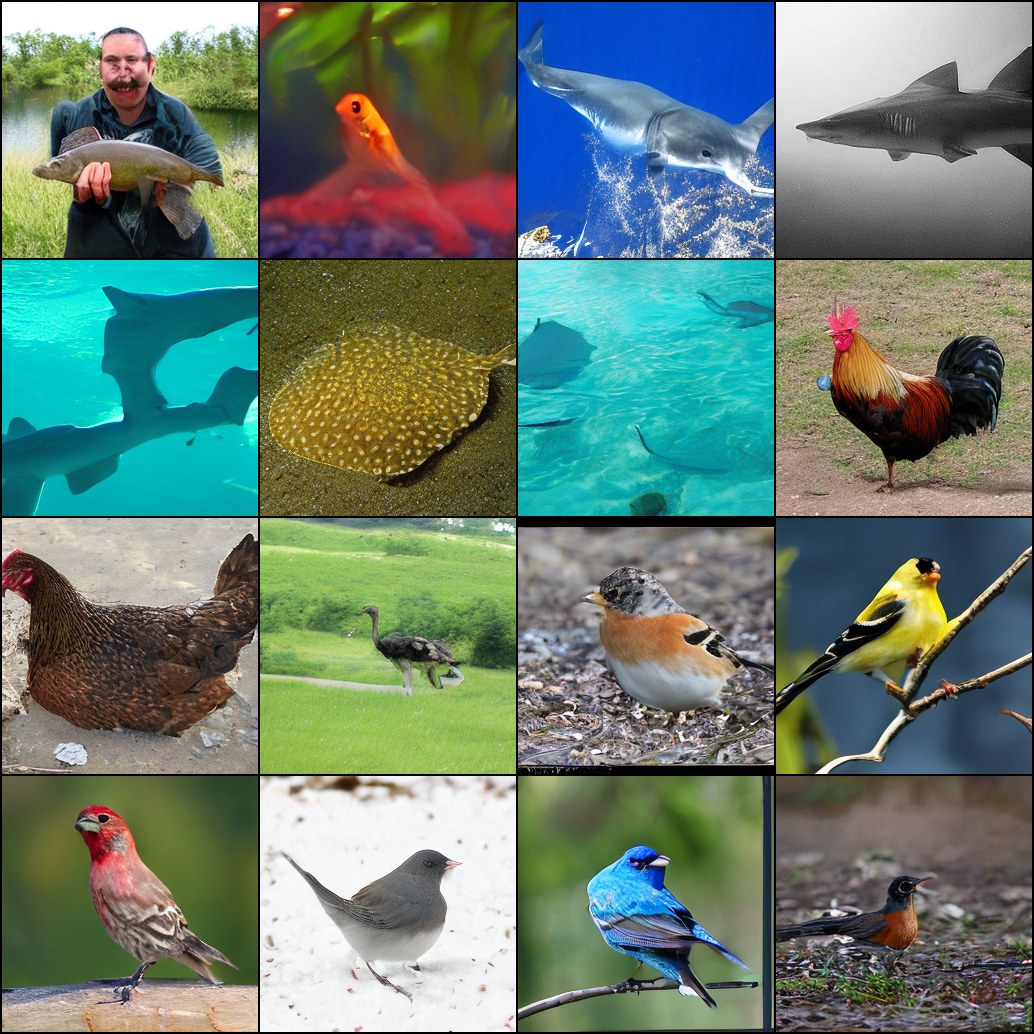} &
\includegraphics[width=.30\linewidth]{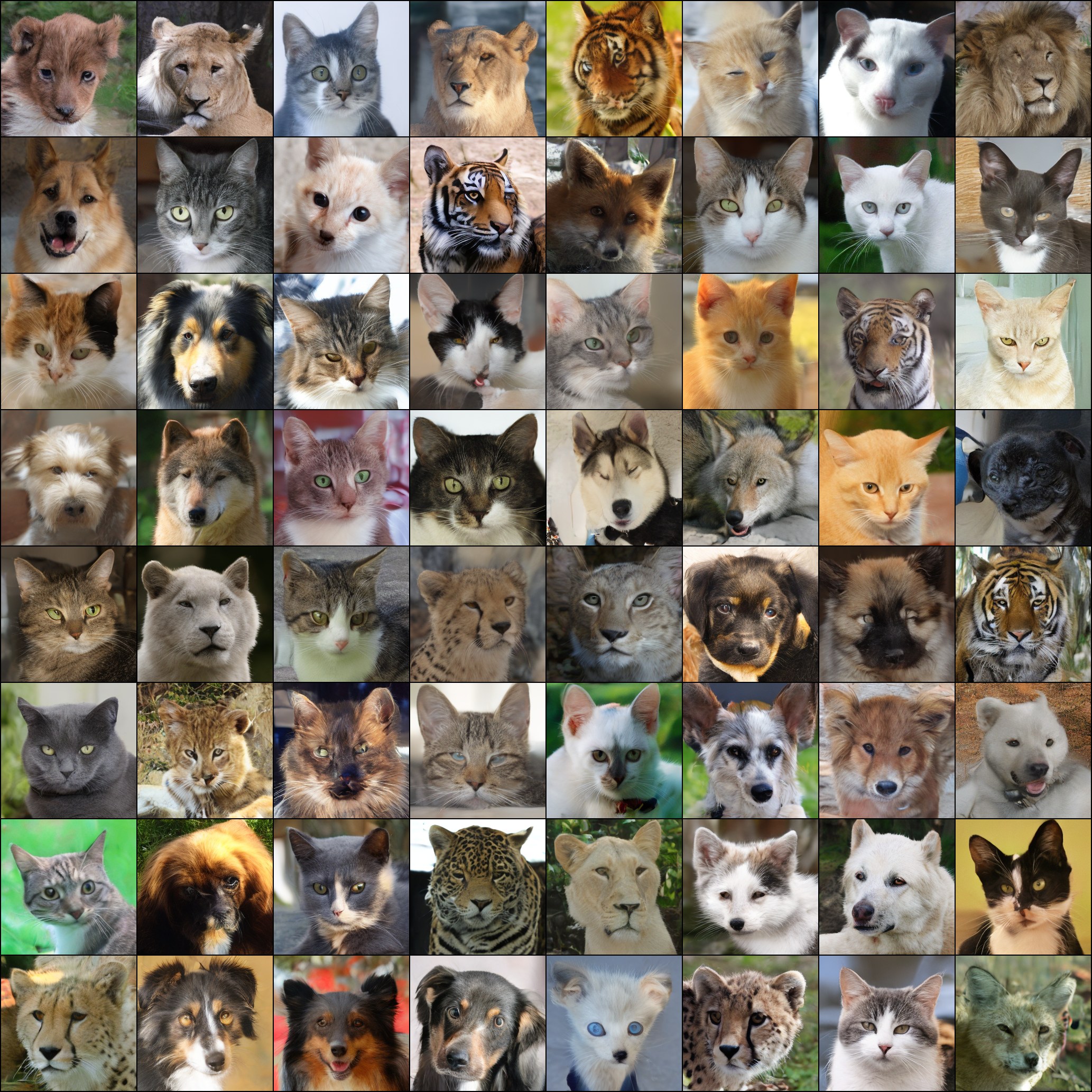} \\
\multicolumn{3}{c}{Local, four velocity queries} \\
\includegraphics[width=.30\linewidth]{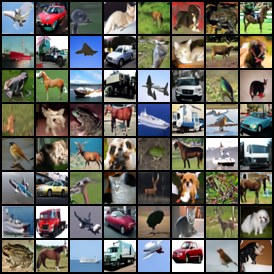} &
\includegraphics[width=.30\linewidth]{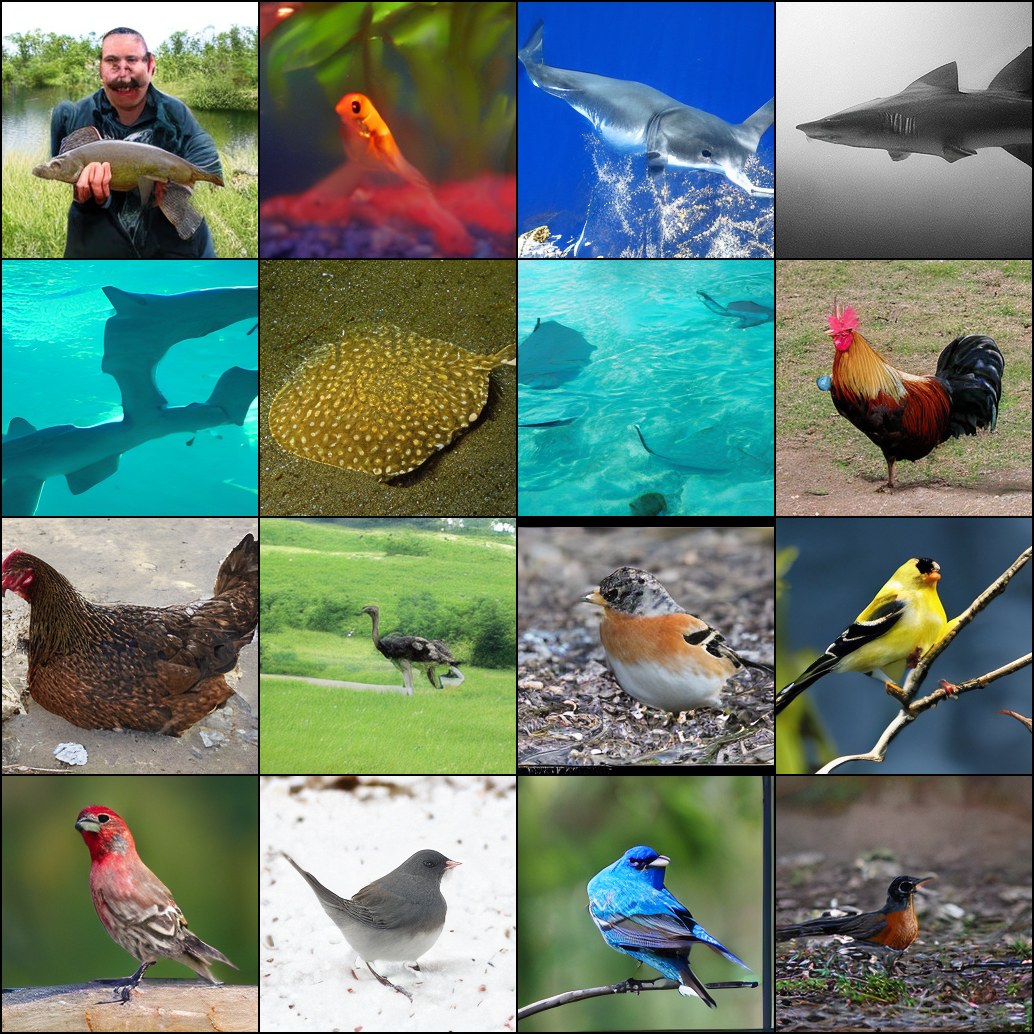} &
\includegraphics[width=.30\linewidth]{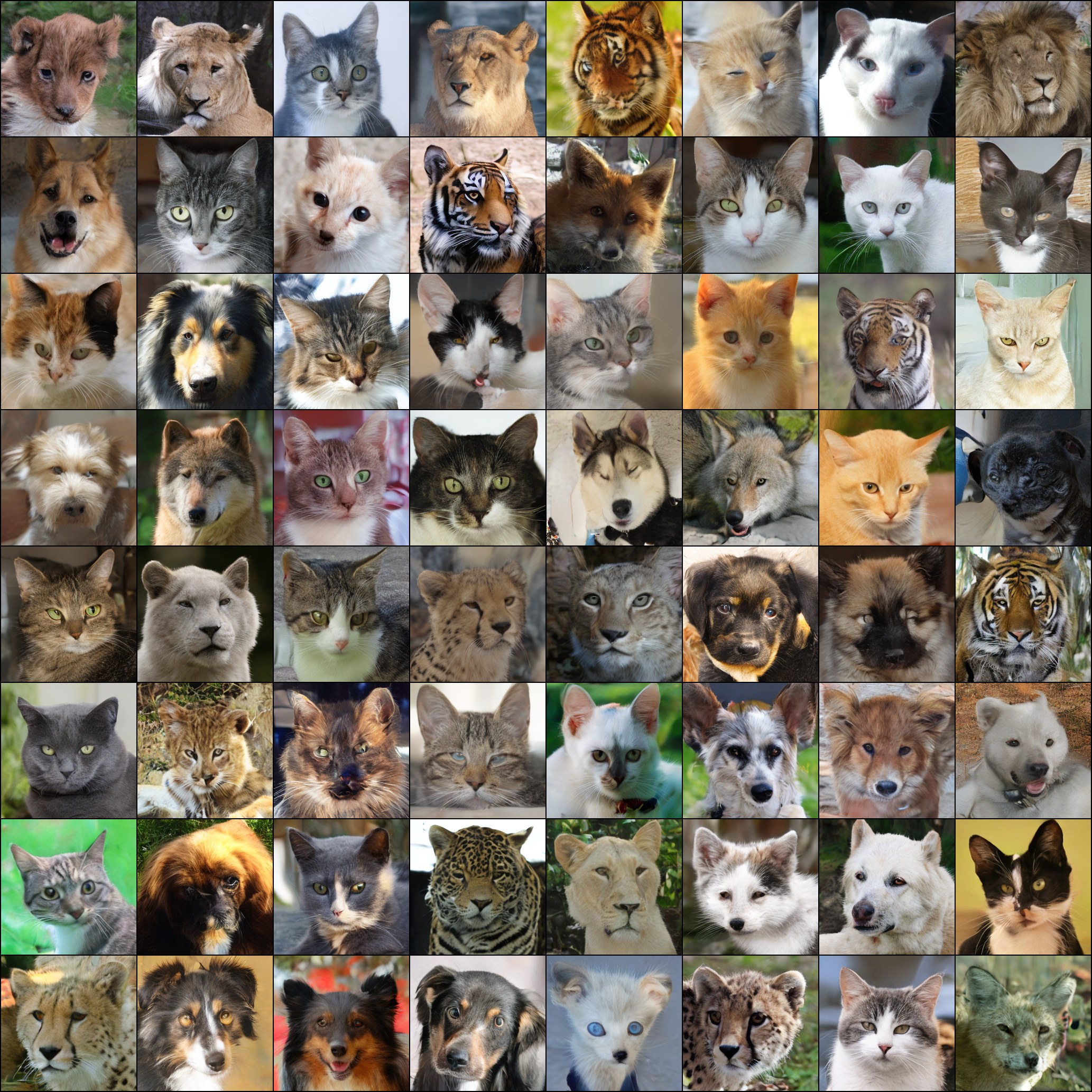} \\
\multicolumn{3}{c}{Horizon, four velocity queries}
\end{tabular}
\caption{Paired, non-curated previews. CIFAR uses checkpoint 300007 and
generation seed 20261001 (64 images per arm); ImageNet uses public MeanFlow-B/2 and
seed 20261061 (16 images per arm); AFHQ uses corrected MeanFlow seed 2027 and generation seed
20261072 (64 images per arm). All use four uniform updates and Local $h=.25$; the Horizon
midpoint is $.50$ for CIFAR, $.30$ for ImageNet and the calibrated $.325$ for
AFHQ. These are different dataset-specific comparisons. Display images use
JPEG compression; their source identities are retained in the figure manifest. Similarity at page scale does not imply identical distributions.}
\label{fig:compact-paired-previews}
\label{fig:qual-cifar}
\label{fig:qual-imagenet}
\label{fig:qual-afhq}
\end{figure}

The previews in Figure~\ref{fig:compact-paired-previews} were emitted by the historical evaluators and compressed as
JPEGs for display; they were not selected by an image-quality score. Within each dataset pair,
cell order, initial noise and, where applicable, conditioning labels coincide. They illustrate paired output
changes and cannot establish the distributional metric claims.

\FloatBarrier

% Compact topic synthesis. Numerical sources are unchanged from the full blind edition:
% temporal_followup.tex; generated_temporal_{schedules,main,budget}.tex;
% generated_endpoint_{main,effects}.tex; historical_transfer_overview.tex;
% reviewed_imagenet_absolute_20260923.tex; generated_supplements_20260921_appendix.tex;
% joint2d_20260922.tex. Full cell tables and frozen reports remain in the archive.
\FloatBarrier
\section{Temporal Controls and Cross-Dataset Transfer}
\label{app:supplements-20260921}
\label{app:historical-transfer-overview}

This section separates equal additional calibration budgets on reused CIFAR checkpoints, transfer of a frozen rule, and target calibration on previously examined public ImageNet models. The final study compares equal-budget candidate allocations that also vary the state grid. All selections precede their corresponding confirmation samples. Later ranges are historically informed; matched new candidate counts do not erase discovery costs or establish continuous family optima. Generation repeats retain the same trained weights.

\subsection{CIFAR: matched additional calibration budgets}
\label{app:temporal-followup}
\label{app:time-controls}

All four families use four updates and four guided velocity evaluations at CFG 1.5, requiring eight conditional/unconditional backbone calls per image. Start changes only $\tau_2$ at $h_2=.25$; Horizon changes only $h_2$ at $\tau_2=.50$; Joint changes both subject to valid query endpoints. These three retain $T=(0,.25,.50,.75,1)$. State-grid search changes the internal update times and ties each query interval to its update interval.

Start has 25 equally spaced candidates on $[.4,.6]$ and Horizon has 25 on $[.25,.5]$. Joint has 25 admissible pairs within those coordinate bounds; state-grid search has 25 schedules. The complete coordinates and ordered lists are in \path{evidence/temporal_candidates.json}. Farthest-point ordering starts from specified historical anchors; ranges, ordering and nested budgets $B\in\{5,13,25\}$ are frozen before development. Each family is charged $B\times10{,}000\times4$ velocity evaluations, including shared schedules. Exact duplicates are generated once, giving 92 distinct development cells. Thus search counts match, but domain geometry and inherited information need not.

The historical MAC-paired seed-42 checkpoint and generation seed 20261501 supply development FID. Each family's lowest-FID candidate is frozen before confirmation seed 20261502 on seven previously evaluated training checkpoints. Each cell uses 10k generated images and the same class-balanced 10k real reference. Table~\ref{tab:temporal-budget} combines the selected coordinates and original 10k results. These Inception measurements use the original truncation path; Appendix~\ref{app:samepixel-20260923} reports the saved-pixel reevaluation.

\begin{table}[ht]
\centering\small
\caption{CIFAR 25-candidate winners and original 10k confirmation means over seven reused checkpoints. P/R use VGG16 features. Other queries are local; state-grid search instead uses $T=(0,.4,.8,.9,1)$. All families have equal additional development counts.}
\label{tab:temporal-budget}\label{tab:temporal-schedules}
\setlength{\tabcolsep}{4pt}
\begin{tabular}{lrrrrrr}
\toprule
Rule & $\tau_2$ & $h_2$ & FID & KID $\times10^3$ & P & R\\
\midrule
Local & .500 & .250000 & 14.204 & 5.811 & .746 & .419\\
State grid & .800 & .100000 & 12.908 & 4.764 & .744 & .438\\
Start & .575 & .250000 & 12.380 & 3.506 & .672 & .506\\
Horizon & .500 & .489583 & 12.902 & 4.901 & .649 & .545\\
Joint & .400 & .500000 & 12.216 & 3.335 & .737 & .465\\
\bottomrule
\end{tabular}
\end{table}

Start beats Horizon in FID on all seven checkpoints (mean margin $\TemporalStartGain$). Joint has lower mean FID than Start by $\TemporalJointGain$, but wins on only $\TemporalJointWins/7$; its descriptive Student-$t$ interval across checkpoints is $[\TemporalJointLow,\TemporalJointHigh]$. It fails the frozen criterion of at least $.1$ mean gain and at least 5/7 wins. Joint also has higher precision and lower recall than Start. The nested-budget comparison uses only checkpoints 300007 and 300017 at all three budgets: mean FID gains over Local at $B=5,13,25$ are $(1.5188,1.5188,1.5165)$ for Horizon, $(1.4280,1.9376,1.9376)$ for Start, $(1.7958,1.7958,1.7958)$ for Joint and $(1.2837,1.2014,1.2014)$ for state-grid search. This subset is not the seven-checkpoint aggregate.

\paragraph{Separate full-reference 50k endpoint.}
\label{app:temporal-50k}
A separately registered confirmation retains Local and the same 25-candidate Start/Horizon winners, with no reselection. Generation seed 20261601 supplies one paired 50k repeat per arm and checkpoint: 21 cells and 1,050,000 images. The real reference is all 50k CIFAR training images. Joint is not included. Main Table~\ref{tab:narrative-calibration}, panel A, reports FIDs $10.272$, $8.311$ and $8.878$ for Local, Start and Horizon, respectively. Horizon improves Local by $1.3942$ FID (descriptive 95\% Student-$t$ interval across checkpoints $[1.2260,1.5625]$); Start improves Horizon by $0.5665$ ($[0.2063,0.9266]$). Both comparisons improve FID and KID on 7/7 checkpoints. Their KID gains are $0.9506$ and $1.2934$ in units of $10^{-3}$. The 10k and 50k rankings agree, but noise, real-reference scope and sample count all change. P/R measured on the 10k images are not 50k measurements, and neither endpoint replaces the primary CFG-2 experiment.

\subsection{Frozen-rule transfer across evaluated checkpoints}

The normalized rule changes only the third query to $h/\delta=1.2$. Table~\ref{tab:cross-data} keeps historical transfer separate from dataset-tuned settings. AFHQ comprises four checkpoints (two training seeds in each of two families), using a latent DiT-B/2 and 10k samples per cell at $256\times256$. The tuned rule uses $H=(.25,.25,.325,.25)$, whereas normalized transfer uses $H=(.25,.25,.30,.25)$. AFHQ development uses seed 42 from each family, with 2k images per candidate, and maximizes the smaller family FID gain after requiring both gains positive; ties use mean gain, then shorter horizon. The selected $.325$ is frozen before the four confirmation checkpoints (seeds 123/2027 in each family) are evaluated. Their KID uses 20 subsets of 500. The tuned $.325$ confirmation gives FID gains of $0.2247$, $0.2371$, $0.1653$ and $0.2192$, but improves KID on only 1/4 checkpoints. The separate normalized $.30$ transfer improves KID on 3/4. These two panels evaluate different frozen rules. Historical CIFAR checkpoints, held-out AFHQ checkpoints and public ImageNet architectures are not interchangeable replication units.

\begin{table}[ht]
\centering\small
\caption{Earlier cross-dataset evidence. Gains are Local minus Horizon. ``Units'' denotes independent training seeds for tuned CIFAR and evaluated checkpoints otherwise; public ImageNet models are not repeated training seeds of one architecture.}
\label{tab:cross-data}
\setlength{\tabcolsep}{4pt}
\begin{tabular}{llcrrr}
\toprule
Rule & Dataset & Units & FID wins & FID gain & KID gain\\
\midrule
Tuned & CIFAR-10 & 7 & 7/7 & \CifarCombinedGain & \CifarCombinedKIDGain\\
Tuned & AFHQv2-256 & 4 & 4/4 & \AFHQTunedGain & mixed\\
$\rho=1.2$ & CIFAR, three recipes & 9 & 9/9 & \CifarCrossFamilyGain & \CifarCrossFamilyKIDGain\\
$\rho=1.2$ & ImageNet-256 & 3 & 3/3 & \PublicImageNetTransferGain & --\\
$\rho=1.2$ & AFHQv2-256 & 4 & 4/4 & \AFHQTransferGain & \AFHQTransferKID\\
\bottomrule
\end{tabular}
\end{table}

\begin{table}[ht]
\centering\small
\caption{Original fixed-rule ImageNet transfer: four uniform updates, third-query $h=.30$ versus $.25$, 50k samples per repeat. Repeats share each public checkpoint's weights. Relative gain divides paired mean FID gain by mean Local FID.}
\label{tab:imagenet-absolute}
\begin{tabular}{lrrrrr}
\toprule
Checkpoint & Repeats & Local FID & Horizon FID & Gain & Relative\\
\midrule
MeanFlow-B/2 & 1 & 5.1859 & 5.0551 & .1308 & 2.52\%\\
MeanFlow-XL/2 & 3 & 2.6300 & 2.5326 & .0974 & 3.70\%\\
AlphaFlow-B/2 & 3 & 5.0633 & 4.9570 & .1063 & 2.10\%\\
\bottomrule
\end{tabular}
\end{table}

Across the three original paired repeats, mean FID gains are $0.09741\pm0.00289$ for MeanFlow-XL/2 and $0.10626\pm0.00569$ for AlphaFlow-B/2 (mean $\pm$ sample SD over generation repeats). All three gains are positive for each checkpoint. B/2 has one original repeat, so it has no within-panel SD. These SDs describe sampling variation at fixed weights, not uncertainty across training seeds. The later confirmation below adds one fresh 50k repeat per checkpoint; the two panels retain their own protocols and are not pooled.

\paragraph{Independent-noise confirmation of the inherited rules.}
\label{app:supp-fixed}
A later panel generates nine new 50k cells at seed 2026092111: Local $(\tau,h)=(.50,.25)$, Horizon $(.50,.30)$ and inherited Start $(.575,.25)$ on the three public models. Table~\ref{tab:supp-fixed-cells} reports every rule without conflating them with target-calibrated winners. Horizon has lower FID than both inherited controls, but lower IS than Local on all three checkpoints. The later panel alone has one repeat per model; the original XL/2 and AlphaFlow repetitions above provide the separate estimates of generation-repeat variability. The gains remain small, and their accompanying IS losses are retained.

\begin{table}[ht]
\centering\small
\caption{Later fixed-rule 50k confirmation. Each entry is FID / IS. These samples are separate from Table~\ref{tab:imagenet-absolute}; none of the three rules is calibrated on these target outputs.}
\label{tab:supp-fixed-cells}
\begin{tabular}{lrrr}
\toprule
Checkpoint & Local & Horizon & Inherited Start\\
\midrule
MeanFlow-B/2 & 5.13619 / 245.217 & 5.00426 / 242.393 & 5.63790 / 211.474\\
MeanFlow-XL/2 & 2.61386 / 288.267 & 2.51254 / 286.031 & 2.87045 / 255.497\\
AlphaFlow-B/2 & 4.95572 / 243.945 & 4.84760 / 241.635 & 5.55436 / 212.154\\
\bottomrule
\end{tabular}
\end{table}

\subsection{ImageNet target calibration and range sensitivity}
\label{app:supp-narrow}
\label{app:supp-range}

All panels retain four uniform state updates and alter only the third query. Native learned guidance stays active. External CFG 1 uses four backbone calls; CFG $.8$ uses eight, so those strata remain separate. Within each model/seed, noise, labels and pre-intervention state hashes are paired. FID uses canonical Inception-2048 and fixed official real moments; IS uses all 1008 detector logits, ten equal splits and a paired seed-plus-100 permutation. The 10k and 50k IS splits contain 1000 and 5000 images, respectively. Split SD is not uncertainty over generation repeats or trained models.

\paragraph{Original six-candidate grids.}
For $d\in\{-.05,-.025,0,.025,.05,.075\}$, Start uses $(.50+d,.25)$ and Horizon uses $(.50,.25+d)$. The Horizon grid therefore includes $h=.20,.225<\delta=.25$ as well as Local and longer intervals. Each family receives six 10k development evaluations at each fixed CFG (1 or $.8$). Shared Local is generated once: 44 development cells, followed by 18 distinct confirmation cells, totaling 620k images. Minimum-FID choices (ties by absolute displacement, then identifier) are frozen after development seed 2026092121 and before confirmation seeds 2026092122/23. At CFG 1, both Horizon choices reach the tested upper endpoint $h=.325$; Start wins in both repeats on both models. At CFG $.8$, B/2 selects Local in both families, while XL/2's small Horizon gain changes sign across repeats ($-.0040,+.0179$).

\paragraph{One specified range check.}
The later, historically informed CFG-1 grids are $\tau\in\{.50,.5125,.525,.5375,.55,.5625\}$ at $h=.25$, and $h\in\{.25,.325,.35,.40,.45,.50\}$ at $\tau=.50$. Each family again receives six 10k evaluations: 22 development and 12 confirmation cells (340k images). Development seed 2026092141 precedes two new confirmation seeds 2026092142/43. Both models select the interior points $\tau=.5375$ and $h=.40$. Start has lower FID and higher IS in all four paired comparisons. Equal counts match new evaluation cost, not perturbation size or search geometry.

\begin{table}[ht]
\centering\footnotesize
\caption{Target-calibration confirmation means over two new 10k repeats. Each score is FID / IS; queries are the frozen winners. Local is always $(.50,.25)$. CFG strata and later ranges remain separate. All repeat-level values and SDs are archived.}
\label{tab:supp-narrow-means}\label{tab:supp-range-means}
\setlength{\tabcolsep}{3pt}
\begin{tabular}{llrrrrr}
\toprule
Grid / external CFG & Model & Start $\tau$ & Horizon $h$ & Local & Start & Horizon\\
\midrule
Narrow / 1 & B/2 & .5250 & .325 & 7.4992 / 190.41 & 7.2680 / 187.38 & 7.3439 / 187.21\\
Narrow / 1 & XL/2 & .5500 & .325 & 5.0265 / 222.08 & 4.8885 / 211.31 & 4.9159 / 219.12\\
Narrow / .8 & B/2 & .5000 & .250 & 7.5811 / 147.15 & 7.5811 / 147.15 & 7.5811 / 147.15\\
Narrow / .8 & XL/2 & .5000 & .225 & 4.9491 / 175.21 & 4.9491 / 175.21 & 4.9421 / 176.38\\
Range / 1 & B/2 & .5375 & .400 & 7.4987 / 191.03 & 7.2297 / 184.14 & 7.3075 / 183.24\\
Range / 1 & XL/2 & .5375 & .400 & 5.0728 / 219.59 & 4.8566 / 214.59 & 4.9277 / 213.02\\
\bottomrule
\end{tabular}
\end{table}

Table~\ref{tab:supp-narrow-means} summarizes the target-calibration confirmations. The original XL/2 CFG-1 panel trades lower Start FID against higher Horizon IS. The range check instead favors Start on both FID and IS, although both Start and Horizon have lower IS than Local. Neither result supports a metric-uniform superiority claim. No confirmation-based reselection or automatic range expansion follows either panel.

\subsection{Fixed composition and two-dimensional calibration}
\label{app:supp-joint}
\label{app:joint2d-20260922}

\paragraph{Fixed composition.}
The first test combines the independently selected Start $\tau=.5375$ and Horizon $h=.40$ without new search. Local, Start, Horizon and Joint receive two fresh paired 10k repeats per model (seeds 2026092151/52; 160k images). Joint has worse FID and IS than Start in all four comparisons (Table~\ref{tab:supp-joint-means}). The mean paired contrasts $\mathrm{FID}_{\rm Start}-\mathrm{FID}_{\rm Joint}$ are $-.81328$ for B/2 and $-.63499$ for XL/2. This tests composition of fixed settings, not optimized two-dimensional calibration.

\begin{table}[ht]
\centering\small
\caption{Fixed composition: two-repeat 10k means, reported as FID / IS. Joint changes both coordinates to $(.5375,.40)$. All four arms use external CFG 1 and four backbone calls.}
\label{tab:supp-joint-means}
\setlength{\tabcolsep}{4pt}
\begin{tabular}{lrrrr}
\toprule
Model & Local & Start & Horizon & Joint\\
\midrule
B/2 & 7.5071 / 188.03 & 7.2339 / 182.83 & 7.3282 / 180.92 & 8.0472 / 166.11\\
XL/2 & 5.0275 / 219.95 & 4.8304 / 214.39 & 4.8879 / 212.36 & 5.4654 / 197.56\\
\bottomrule
\end{tabular}
\end{table}

\paragraph{Frozen nine-point calibration.}
A separate study evaluates the Cartesian grid $\tau\in\{.5000,.5250,.5375\}$, $h\in\{.25,.30,.40\}$ on the same two public checkpoints, at external CFG 1 and four calls. It contains Local, the prior Start, the prior Horizon and the failed fixed Joint. Each model receives nine paired 10k development cells (seed 2026092271; 180k images across both models). The finite minimum-FID winner is sealed before two fresh 10k confirmation seeds, 2026092272/73. Exact numerical ties use changed-axis count, absolute perturbation, $\tau$, $h$, then candidate key. This is additional joint calibration against a fixed prior Start, not equal-new-budget optimization of all families.

Both models select Start $(.5375,.25)$ (development FID $7.072$ and $4.784$). Confirmation retains Local, Start, Horizon and the winner role, but identical roles share one physical cell: \JointTwoDConfirmCells{} cells and \JointTwoDConfirmImages{} images, or \JointTwoDImages{} formal images including development. Winner-minus-Start FID is exactly zero and IS retention exactly 100\% by identity, not an independent estimate of a null effect. The frozen requirement---strictly lower FID in all four repeats, mean gain at least $.05$ per model, and at least 99\% IS retention per repeat---therefore fails. No 50k expansion follows. This finite grid supplies no demonstrated increment beyond Start; it does not exclude useful points outside the grid.

\paragraph{Numerical and interpretive scope.}
Independent audits recompute stored-feature FID and logit-based IS, check selections, query counts, paired stream hashes and source identities. Earlier panels retain stream digests rather than original noise/midpoint tensors; the joint-grid panel additionally saves the canonical metric pixels. Matched real per-image ImageNet features are unavailable, so these studies do not report KID or manifold P/R; the 1008 detector logits are not treated as accuracy against 1000 conditioning labels. Complete candidates, repeat-level cells, numerical provenance and audits remain in the evidence archive. These fixed-model studies establish specific metric orderings under finite budgets, not global optimality or a rule for unseen checkpoints.
\FloatBarrier

\subsection{Equal-budget public grid/start/horizon allocations}
\label{app:public-grid-increment}
\paragraph{Frozen comparison.}
On the previously evaluated MeanFlow-B/2 and XL/2 public weights, we compare two finite allocations: Grid+Start (GS) and Grid+Start+Horizon (GSH). Both use four state updates and four backbone calls, with native learned guidance and external CFG~1; external CFG~1 does not make these learned-guidance checkpoints unguided. Each allocation contains 12 candidates: shared uniform Local and Start $(\tau_2,h_2)=(.5375,.25)$ anchors, and ten matched grid/start points from one frozen scrambled Sobol construction (seed 2026092681). Grid bounds are $t_1\in[.20,.30]$, $t_2\in[.45,.55]$, $t_3\in[.70,.80]$. Only the third query start has an offset in $[-.05,.05]$. At that update, GS uses $h_2=\delta_2$, while GSH uses $h_2/\delta_2\in[.5,1.5]$, including shorter and longer horizons. Illegal paired queries are rejected without clipping. The two allocations share coordinates but GSH is not a strict superset of the GS candidates.

\paragraph{Budget and selection.}
Each candidate first receives 2k paired development samples; the two lowest-FID candidates in each model/allocation receive 8k additional samples. Selection then uses FID on the combined original 2k and appended 8k pool, with frozen candidate order breaking exact ties. This charges $12\times2\mathrm{k}+2\times8\mathrm{k}=40\mathrm{k}$ images per model/allocation. Shared candidates are generated once and savings are not reassigned. The frozen GS/GSH winners and the GSH winner with $h_2$ restored to $\delta_2$ receive two fresh paired 10k repeats. Actual counts are 88k screening, 56k top-up and 100k confirmation images: 244k total (61 physical cells), below the 280k ceiling. Noise and labels are paired across candidates by global sample index; intermediate states on different grids need not match.

\paragraph{Decision and outcome.}
Let $A=\mathrm{FID}(\mathrm{GS})-\mathrm{FID}(\mathrm{GSH})$ and $B=\mathrm{FID}(\mathrm{restored})-\mathrm{FID}(\mathrm{GSH})$. The prespecified descriptive gate requires both contrasts positive in each repeat, both mean gains at least .05 for each model, and GSH retaining at least 99\% of each reference IS in each repeat. This is not a significance test. Both models fail the FID conditions; every IS retention condition passes. B/2 selects \texttt{gs\_08} and \texttt{gsh\_07} (restored: \texttt{gs\_07}); XL/2 selects \texttt{gs\_08} and the Start anchor, whose restoration is identical. For B/2, mean $A=-0.02712$ and $B=+0.02447$; for XL/2, $A=-0.03583$ and $B=0.00000$. Table~\ref{tab:public-grid-repeats} reports both confirmation repeats. The small positive B/2 intervention effect does not produce a better selected allocation. All registered confirmations were completed, and the study stopped without an automatic 50k expansion or rescue search.

\begin{table}[htbp]
\centering\small
\setlength{\tabcolsep}{4pt}
\begin{tabular}{llrrrrrr}
\toprule
Model & Repeat & GS FID & GSH FID & Restored FID & $A$ & $B$ & Min. IS ratio\\
\midrule
B/2 & 1 & 7.2216 & 7.2564 & 7.2808 & -0.0347 & +0.0244 & 0.9941\\
B/2 & 2 & 7.2238 & 7.2433 & 7.2678 & -0.0195 & +0.0245 & 0.9956\\
XL/2 & 1 & 4.8049 & 4.8307 & 4.8307 & -0.0258 & +0.0000 & 1.0000\\
XL/2 & 2 & 4.7888 & 4.8347 & 4.8347 & -0.0459 & +0.0000 & 0.9949\\
\bottomrule
\end{tabular}
\caption{Both fresh 10k confirmation repeats. $A=\mathrm{FID}(\mathrm{GS})-\mathrm{FID}(\mathrm{GSH})$ compares selected allocations; $B=\mathrm{FID}(\mathrm{restored})-\mathrm{FID}(\mathrm{GSH})$ isolates horizon at the selected grid/start. Positive favors GSH. The last column is the smaller of the GSH/GS and GSH/restored IS ratios. Repeats use seeds 2026092692 and 2026092693; they are sampling repeats on fixed public weights.}
\label{tab:public-grid-repeats}
\end{table}

\paragraph{Scope and reconstruction.}
These are two reused public checkpoints and one historically informed search construction, not independent training replications or search-seed replications. The 2k screening FID uses 2048-dimensional features and can discard useful candidates; this finite comparison establishes neither continuous-family optima nor general dominance. FID and IS do not resolve precision--recall trade-offs. The evidence archive includes every candidate, metric report, seal, audit and runtime scheduling amendment. The portable checker reconstructs selection, contrasts, gates and counts from those reports; it does not re-extract features or recompute image metrics. The original independent audit checked metrics from retained features and artifact hashes without repeating Inception extraction. Image arrays, feature tensors and model weights are outside the portable package.

% Compact topic synthesis of unchanged frozen measurements:
% samepixel_appendix_20260923.tex (contrasts and quantization table),
% dino_seven_20260922.tex (DINO contrasts), multimetric_scope_20260922.tex,
% guidance_evidence.tex (earlier two-checkpoint representation audit).
% Per-cell reports, original tables and all evaluator records remain archived.
\FloatBarrier
\section{Metric Robustness and Quality--Coverage Trade-offs}
\label{app:samepixel-20260923}
\label{app:multimetric-scope-20260922}

The seven-checkpoint representation audit and same-pixel Inception reevaluation test whether metric rankings survive a change of evaluator on fixed samples. Both reuse saved samples without new training, generation or schedule search. Correlated metrics are not independent replications.

\subsection{Fixed samples and evaluator definitions}
\label{app:dino-seven-20260922}
\label{app:dino-audit}

The main audit uses all seven reused training checkpoints in the 25-candidate CIFAR temporal confirmation, with Local, Start and Horizon fixed at CFG 1.5. Their third queries are $(\tau_2,h_2)=(.5,.25)$, $(.575,.25)$ and $(.5,.4895833333)$. Each cell retains all 10k images from generation seed 20261502: 210,000 images total. The real reference is the same class-balanced set of 10k CIFAR training images. This is a post-hoc measurement sensitivity study of the 10k calibration panel, distinct from the primary CFG-2 rule and separate 50k endpoint.

DINOv2 ViT-S/14~\citep{oquab2023dinov2} uses 384-dimensional final-LayerNorm CLS features, without L2 normalization, PCA or learned calibration. The frozen official implementation applies PIL bicubic resize to 256, center crop to 224 and ImageNet normalization to real and generated images; its source and weights are hash-bound in the archive. Upsampling and cropping are part of the evaluator. DINO distance is the squared Gaussian Wasserstein distance using float64 empirical moments; manifold precision/recall uses $k=3$ on raw features. These coordinates describe coverage and concentration in the chosen representation, not a single quality score.

The original Inception pipeline truncates clipped floating outputs after multiplication by 255. DINO and the original VGG16 \texttt{fc2\_relu} P/R pipeline instead consume saved rounded uint8 images. A channel can differ by one uint8 level. To remove that input mismatch, the same-pixel audit passes the saved uint8 tensors directly to the unchanged TorchMetrics/torch-fidelity Inception2015 extractor with \texttt{normalize=False}. The real indices and pixel bytes also match. This produces 2048-dimensional features; FID retains float64 moments and unbiased sample covariance. KID retains the unbiased squared-MMD estimator with a degree-3 polynomial kernel, 100 subsets of 1000 and CPU permutation seed 20261602. No additional rounding or truncation is applied.

Inception's native resizing and DINO's resize/crop remain different, so the audit compares complete evaluators rather than isolating feature weights. Every checkpoint contributes one paired difference to each descriptive Student-$t$ interval. All three arm contrasts are reported without multiplicity-adjusted significance claims or outcome-based reselection.

\subsection{Ranking reversal survives identical image bytes}

Main Table~\ref{tab:narrative-calibration}, panel B, reports the three same-pixel operating points. Table~\ref{tab:samepixel-contrasts} gives their paired Inception changes; Table~\ref{tab:dino-seven-contrasts} gives the DINO changes on exactly those saved images. Horizon lowers FID, KID and DINO distance versus Local on all seven checkpoints, while its DINO precision decreases and recall increases on every checkpoint. Start has lower FID/KID than Horizon on 7/7, but Horizon has lower DINO distance and higher DINO precision/recall than Start on 7/7. Start's DINO distance is worse than Local on all seven despite its lower FID/KID. Pixel quantization mismatch is therefore not required for this ordering reversal; universal perceptual superiority does not follow.

\begin{table}[ht]
\centering\footnotesize
\caption{Inception contrasts after reevaluation on the saved DINO input bytes. Changes are first arm minus second; negative values favor the first arm. Intervals are descriptive 95\% Student-$t$ intervals across checkpoints. Both KID columns are scaled by $10^3$.}
\label{tab:samepixel-contrasts}
\setlength{\tabcolsep}{3pt}
\begin{tabular}{lrrcrrc}
\toprule
Contrast & $\Delta$FID & 95\% interval & Lower & $\Delta$KID & 95\% interval & Lower\\
\midrule
Horizon -- Local & -1.298 & $[-1.472,-1.123]$ & 7/7 & -0.898 & $[-1.064,-0.732]$ & 7/7\\
Start -- Local & -1.824 & $[-2.130,-1.517]$ & 7/7 & -2.296 & $[-2.714,-1.879]$ & 7/7\\
Horizon -- Start & +0.526 & $[+0.206,+0.846]$ & 0/7 & +1.398 & $[+0.900,+1.896]$ & 0/7\\
\bottomrule
\end{tabular}
\end{table}

\begin{table}[ht]
\centering\small
\caption{DINO contrasts on the same 21 saved image cells. Changes are first arm minus second. Distance decreases are preferable; P/R changes are percentage points (pp). The distance intervals describe variation over the seven reused checkpoints. P/R are separate coordinates.}
\label{tab:dino-seven-contrasts}
\setlength{\tabcolsep}{4pt}
\begin{tabular}{lrrcrr}
\toprule
Contrast & $\Delta$distance & 95\% interval & Lower & $\Delta$P (pp) & $\Delta$R (pp)\\
\midrule
Horizon -- Local & -3.132 & $[-4.104,-2.160]$ & 7/7 & -2.04 & +3.05\\
Start -- Local & +14.690 & $[+11.789,+17.591]$ & 0/7 & -3.50 & +0.33\\
Horizon -- Start & -17.822 & $[-19.846,-15.798]$ & 7/7 & +1.46 & +2.72\\
\bottomrule
\end{tabular}
\end{table}

The original VGG P/R means are $(.7464,.4188)$ for Local, $(.6722,.5057)$ for Start and $(.6494,.5446)$ for Horizon. Thus the original manifold evaluator also shows lower Horizon precision and higher recall versus Local, and even the ordering of precision between Start and Horizon depends on the feature space. Those VGG values remain measurements of the original saved 10k cells, not of the 50k confirmation images.

\subsection{Quantization sensitivity and retained historical values}

Table~\ref{tab:samepixel-quantization} retains the original Inception results beside the saved-pixel reevaluation. The changes are small relative to the arm differences and preserve every Start/Horizon FID/KID ordering. Original floating endpoints were not retained, so the changed-pixel fraction and a separate DINO effect of quantization cannot be reconstructed. The new measurements do not overwrite the historical reports.

\begin{table}[ht]
\centering\footnotesize
\caption{Rounded-input reevaluation versus original truncation-input Inception measurements. Changes are new minus original. KID columns are scaled by $10^3$. All checkpoint cells and KID sensitivity intervals remain in the archive.}
\label{tab:samepixel-quantization}
\setlength{\tabcolsep}{3pt}
\begin{tabular}{lrrrrrrr}
\toprule
Arm & Old FID & New FID & $\Delta$FID & 95\% interval & Old KID & New KID & $\Delta$KID\\
\midrule
Local & 14.204 & 14.212 & +.0074 & $[+.0025,+.0124]$ & 5.8109 & 5.8106 & -.0003\\
Start & 12.380 & 12.388 & +.0078 & $[+.0022,+.0134]$ & 3.5057 & 3.5144 & +.0087\\
Horizon & 12.902 & 12.914 & +.0121 & $[+.0076,+.0166]$ & 4.9006 & 4.9125 & +.0119\\
\bottomrule
\end{tabular}
\end{table}

An earlier, separate DINO audit used two checkpoints and three inherited rules at four CFG values (24 saved 10k cells). For the inherited rules at CFG 1.5, Horizon lowered DINO distance versus Local by 3.935 on average, while precision fell by 2.46 pp and recall rose by 3.63 pp. Horizon beat Start on DINO distance, precision and recall at all eight checkpoint/CFG operating points, but its DINO distance was worse than Local at CFG 1 on both checkpoints. Those earlier rules differ from the seven-checkpoint development-selected rules above; their cell histories are not pooled.

\subsection{Coverage limits and independent verification}

The seven-checkpoint audit supplies DINO coverage for the fixed CFG-1.5 10k cells only. It does not measure the old CFG-2 primary rule, old 50k endpoints, six cross-recipe checkpoints, ImageNet or AFHQ. The new blind NFE study has its own separately frozen multi-metric scope (Appendix~\ref{app:blind-nfe-20260925}); it is not an extension of these old saved-image samples. In the separate cross-recipe prefix panel, all 12 Horizon--Local comparisons reduce VGG precision and requested-label classifier agreement while increasing recall. Agreement is the fraction of samples whose top-1 prediction from the frozen public CIFAR ResNet20 matches the requested class. The classifier achieves 92.59\% accuracy on the real 10k CIFAR test set; its source revision, weights and preprocessing are recorded in \path{evidence/supplements_20260921/prefix/protocol.json}. Label agreement is not a human perceptual judgment.

ImageNet fixed transfer reduces IS while improving FID; target-calibration ordering depends on the tested grid and CFG (Appendix~\ref{app:supplements-20260921}). IS does not measure manifold precision or recall. Matching real per-image ImageNet features are unavailable, so no KID or P/R is inferred from the stored real moments. AFHQ KID improves on 1/4 checkpoints in the tuned $.325$ confirmation and on 3/4 in the separate normalized $.30$ transfer. No human perceptual study or cross-metric success prediction is claimed.

Independent audits bind input identities and recompute all 21 DINO distances, 42 DINO P/R coordinates and 21 same-pixel FID/KID cells from retained features. The same-pixel reevaluation re-extracts features from 210k generated and 10k real images; the independent extraction check covers a 2816-image subset, not the full image set. Both audits add zero generated images. Full reports, protocols, numerical provenance, certificates and tensor manifests remain in \path{evidence/dino_seven_20260922/} and \path{evidence/samepixel_20260923/}; large tensors and weights are outside the manuscript ZIP. No outcome triggers rescue search.
\FloatBarrier

% Compact editorial view of unchanged evidence/guidance, supplements_20260921/prefix,
% and evidence/mechanism_20260924/numeric_ledger.json. Full cell tables remain
% in sections/mechanism_appendix_20260924.tex and archive/full_manuscript_20260925.pdf.
\section{Position response, supporting controls and directional diagnostics}
\label{app:prefix-mechanism}
\label{app:guidance-evidence}

\subsection{Position and prospective guidance-history replication}
The single-position scan (Table~\ref{tab:time-horizon-seven}) holds the four-update grid fixed at CFG 2 and tests $h\geq\delta=.25$. At the start, every tested non-local horizon worsens FID on all seven checkpoints; at the midpoint, each tested non-local horizon improves it on all seven. For $h=.50$ at $t=.50$, mean FID gain is 2.010. This 10k position scan is separate from the 50k frozen-rule endpoint. The prefix control tests whether query time alone describes this response; it does not identify an error mechanism.

The guidance-prefix factorial uses the uniform grid $(0,.25,.5,.75,1)$, CFG schedule $(p,p,1.5,1.5)$ with $p\in\{1,1.5\}$, and midpoint horizon $h_2\in\{.25,.50\}$. All arms use four updates and eight conditional/unconditional backbone calls, including when $p=1$. Local and Horizon share byte-identical entering states within each prefix. Changing the first two guidance values changes that state while keeping guidance at and after the intervention fixed. Define
\begin{equation}
G_p=\mathrm{FID}(p,.25)-\mathrm{FID}(p,.50),\qquad D=G_{1.5}-G_1.
\end{equation}
The pilot uses checkpoints 300007/300017 and two fresh generation repeats; the subsequent replication fixes the settings and directional hypothesis before evaluating the five remaining checkpoints on new noise. Those five weights already appeared in other horizon studies, but their factorial outcomes did not select this hypothesis. The preregistered replication gate requires all five $D>0$ and their mean above .1 FID. It passes: mean $D=\PrefixFiveGain$, descriptive 95\% Student-$t$ interval across checkpoints $[\PrefixFiveLow,\PrefixFiveHigh]$, with positive KID contrasts on all five. Table~\ref{tab:prefix-contrasts} reports the checkpoint contrasts. Sampling repeats are averaged within checkpoint, not treated as training replications.

% Generated from hash-verified JSON; do not edit numeric cells.
\begin{table}[ht]
\centering
\small
\caption{Guidance-prefix contrasts. $G_p$ is local-minus-horizon FID after prefix CFG $p$, with both suffix steps fixed at CFG $1.5$. Pilot repeats are averaged within checkpoint. The replication hypothesis and settings were frozen after the two-checkpoint pilot and before the five new factorial evaluations.}
\label{tab:prefix-contrasts}
\begin{tabular}{lrrrrr}
\toprule
Cohort & Train seed & $G_1$ & $G_{1.5}$ & $D=G_{1.5}-G_1$ & $D_{\rm KID}\times10^3$ \\
\midrule
Pilot & 300007 & 0.1992 & 1.5668 & 1.3676 & 1.3368 \\
Pilot & 300017 & 0.4131 & 1.6243 & 1.2112 & 1.1190 \\
Replication & 700001 & 0.1046 & 1.2689 & 1.1642 & 1.1723 \\
Replication & 700027 & 0.1701 & 1.2305 & 1.0604 & 1.1565 \\
Replication & 700057 & -0.1728 & 1.1677 & 1.3405 & 1.3043 \\
Replication & 700067 & 0.0941 & 1.2851 & 1.1910 & 1.1661 \\
Replication & 700079 & 0.3489 & 1.5875 & 1.2385 & 1.2969 \\
\bottomrule
\end{tabular}
\end{table}

\paragraph{Related diagnostics and scope.}
\label{app:component-mechanism}
An earlier conditional/guidance decomposition changes one term at a time in $u^c+(w-1)(u^c-u^n)$ at $w=1.5$. Conditional-only replacement yields mean FID gain \AttributionConditional{}, compared with \AttributionFull{} for Full; guidance-vector-only replacement gives \AttributionGuidance{} despite slightly improving KID. These constructed arms cost ten backbone calls, versus eight for native sampling. They do not isolate an unknown true-flow error. The subsequent norm/direction diagnostics below use different fixed interventions.

\label{app:supp-prefix}
A separate cross-recipe extension uses three historical seeds per MAC-paired and semidiscrete recipe. Mean FID prefix interactions are \SuppPrefixOfficialDID{} and \SuppPrefixSemidiscreteDID{}, respectively. Across its twelve horizon contrasts, precision decreases, recall increases and classifier agreement slightly decreases. Semidiscrete seed 123 has worse FID at prefix 1; MAC-paired seed 42 has slightly worse KID there. This is historical-recipe evidence, not six new training runs. ImageNet prefix transfer has inconsistent repeat signs (Appendix~\ref{app:public-prefix-followup}).

% Generated from three unchanged historical CFG-1 reports; no new images.
% Reproduce/check: python repro/build_existing_cfg_scope_20260926.py --check
\subsection{Guidance and shorter-horizon coverage}
\label{app:horizon-scope}
Existing CIFAR runs also set CFG to 1 throughout the trajectory: this is
class-conditional sampling without external CFG extrapolation, not unconditional
sampling. They retain four uniform updates and compare midpoint $h=.50$ against
Local $h=.25$. FID changes are small and mixed; KID worsens on all three
checkpoints (Table~\ref{tab:existing-cfg1-scope}). These historical runs do not
establish a position reversal without external CFG. VGG precision decreases and
recall increases on all three.

\begin{table}[htbp]
\centering\footnotesize
\setlength{\tabcolsep}{4pt}
\begin{tabular}{lrrrr}
\toprule
& \multicolumn{2}{c}{FID $\downarrow$} & \multicolumn{2}{c}{KID$\times10^3\downarrow$}\\
Checkpoint & Local & Horizon & Local & Horizon\\
\midrule
Local MAC s42 (dev.) & 16.7540 & 16.7410 & 10.5848 & 10.9653 \\
Paired recipe s300007 & 15.9024 & 16.0348 & 9.4296 & 9.9692 \\
Paired recipe s300017 & 15.7868 & 15.6920 & 9.3806 & 9.6567 \\
\bottomrule
\end{tabular}
\caption{Existing full-trajectory CFG-1 results: one paired 10k generation repeat
per checkpoint against a fixed class-balanced 10k real subset. All weights were
trained locally; the MAC baseline is not a released checkpoint.
Seed 42 was development; the paired-recipe seeds reuse historical
weights. Inception uses the historical float-input \texttt{normalize=True} path,
not the rounded-uint8 same-pixel protocol of Appendix~\ref{app:samepixel-20260923}.
The historical sampler still executes eight backbone calls.}
\label{tab:existing-cfg1-scope}
\end{table}

\paragraph{Existing shorter-horizon coverage.}
The blind CIFAR NFE~4/8 candidate sets already include $h=.8\delta$ at CFG 1.5
(Appendix~\ref{app:blind-nfe-20260925}); these are 5k calibration evaluations,
without independent confirmation of that short candidate. The original
ImageNet six-candidate grid includes $h=.20,.225$ around Local $h=.25$
(Appendix~\ref{app:supp-narrow}). Native learned guidance remains active there,
including at external CFG 1. These panels cover short horizons in their stated
settings, not a complete bidirectional position scan without external CFG.

\subsection{Four sequential directional studies}
\label{app:mechanism-20260924}
\label{app:velocity-factorization}
The projection, equal-norm, signed-dose and prefix panels (Table~\ref{tab:mechanism-design}) were designed sequentially: later questions used earlier results, but each panel froze its models, arms, inputs, metrics and stopping rules before its own new generation. All six distinct weights had appeared in earlier studies. The prefix panel reuses the signed pair, so the sequence is exploratory and does not form a pooled six-model confirmatory test.

\begin{table}[htbp]
\centering\small
\begin{tabular}{llrr}
\toprule
Panel & Existing training seeds & Repeats/model & Images \\
\midrule
Projection & 300007, 300017 & 2 & 160,000 \\
Equal norm & 700001, 700027 & 1 & 100,000 \\
Signed dose & 700057, 700067 & 1 & 100,000 \\
Prefix & 700057, 700067 & 1 & 120,000 \\
\bottomrule
\end{tabular}
\caption{Directional panels: 48 unique 10k cells, ten checkpoint--generation jobs, six existing weights and no new training. Every arm is retained in the portable numeric ledger.}
\label{tab:mechanism-design}
\end{table}

\paragraph{Shared definitions.}
Within each fixed-prefix comparison, the arms differ only in the third velocity on the uniform four-update grid, with physical duration $\delta=.25$. The prefix panel additionally varies earlier guidance as specified below. Unless explicitly varied, CFG is 1.5. At a shared midpoint state $x$, let $v(t,h,w)$ denote the CFG-combined velocity and set
\begin{align}
v_0&=v(.5,.25,1.5),&v_H&=v(.5,.4895833333,1.5),\nonumber\\
d&=v_H-v_0,&s&=v(.5125,.25,1.5)-v_0,\label{eq:mechanism-secants}\\
g&=.1\,[u_c(x,.5,.25)-u_n(x,.5,.25)],&p&=Pd,\quad r=d-p.\nonumber
\end{align}
Here $P$ projects onto the retained span of $s,g$ after flattening CHW coordinates. Differences and projection use float64: normalize nonzero columns above $10^{-12}$, symmetrize their Gram matrix, and retain eigenvalues strictly above $10^{-10}\lambda_{\max}$. Rank zero gives $P=0$; no clipping, ridge search or endpoint-dependent fit is used. Corrections are converted separately to float32. Local and Full retain their native float32 arithmetic. The horizon differs slightly from the earlier prefix experiment's $.50$.

All arms within a prefix share initial noise, requested labels and the entering state; subsequent trajectories evolve separately. Inception FID/KID and DINO distance/precision/recall consume the same rounded uint8 pixels against the same frozen class-balanced 10k real reference. Feature extraction is float32; covariance calculations use float64. KID uses 100 subsets of 1,000; DINO uses unnormalized ViT-S/14 CLS features and three-neighbor precision/recall. Means first aggregate repeats within checkpoint and then give the two checkpoints equal weight. Metrics from a panel are correlated outcomes, not independent replications.

\paragraph{Projection.}
The four velocities are $v_0$, $v_0+p$, $v_0+r$ and $v_H$. Full must first improve both FID and DINO over Local in all four checkpoint--generation-repeat pairs. Both primary contrasts, Residual/Local and Full/Projected, then require positive gains in both distances in every pair and a mean FID gain of at least .05 within each checkpoint. These conditions pass; Table~\ref{tab:revised-mechanism-gate-minima} gives the weakest pair and checkpoint margins. The residual retains approximately 67--69\% of correction energy, a geometric fraction rather than a percentage of endpoint benefit explained. Projected improves FID but worsens DINO; Full improves KID over Projected in only two of four pairs. Residual's mean precision falls .0100 relative to Local while recall rises .0517.

\paragraph{Equal norm and sign controls.}
The norm panel compares Residual with Local, Full, scaled Full $v_0+(\|r\|/\|d\|)d$, and Random $v_0+(\|r\|/\|q\|)q$, where $q=(I-P)z$ is a projected Gaussian vector. The Gaussian generator is fixed independently of initial noise. Zero residual gives zero control corrections; invalid small denominators raise an error instead of triggering resampling. Each primary Residual comparison against Local, scaled Full and Random requires positive FID/DINO gains and at least .05 FID on each checkpoint; all pass. Scaled Full reduces FID and DINO distance relative to Full by .105 and 2.676, respectively, and Random reduces them relative to Local by .104 and 1.129, so attenuation and random perturbations can help but do not match Residual here. These controls do not exclude every optimized scalar schedule or random-direction distribution.

The signed panel fixes $v_0+\alpha r$ for $\alpha\in\{-1,-.5,0,.5,1\}$. Negative corrections are exact negations of their positive float32 counterparts: opposite signs therefore match correction norm and Fourier power, though not total velocity or endpoint displacement; their alignment with the baseline velocity remains uncontrolled. The separately frozen $+r/-r$ and $+r$/Local gates require positive FID/DINO gains and at least .05 FID on each model; both pass. FID, KID and DINO decrease across the five ordered doses on each model (Table~\ref{tab:revised-signed-doses}). This observed finite response is neither an optimal-dose estimate nor a preregistered monotonicity test. Both negative doses perform worse than Local; $+r$ lowers mean precision by .00975 and raises recall by .05020 relative to Local.

\begin{table}[htbp]
\centering\small\setlength{\tabcolsep}{4pt}
\begin{tabular}{llrr}
\toprule
Panel & Treatment / reference & FID gain & DINO gain \\
\midrule
Projection & Residual / Local & +2.817 & +16.560 \\
 & Full / Projected & +0.850 & +10.077 \\
 & Projected / Local & +0.745 & -5.899 \\
Equal norm & Residual / scaled Full & +1.165 & +9.728 \\
 & Residual / Random & +2.352 & +14.456 \\
Signed & $+r$ / $-r$ & +7.158 & +78.923 \\
 & $+r$ / Local & +2.314 & +16.528 \\
 & $-.5r$ / Local & -1.870 & -23.061 \\
 & $-r$ / Local & -4.844 & -62.395 \\
Prefix 1 & Full / Local & -0.003 & -9.112 \\
 & Residual / Local & +1.973 & +11.725 \\
Prefix 1.5 & Full / Local & +1.187 & +2.563 \\
 & Residual / Local & +2.350 & +16.798 \\
\bottomrule
\end{tabular}
\caption{Selected contrasts retain the adverse controls. Gains are reference minus treatment; positive favors the treatment. These checkpoint means summarize the sequential panels, not an equal-cost sampler comparison. Tables~\ref{tab:revised-mechanism-gate-minima} and~\ref{tab:revised-signed-doses} give all gate minima and signed doses; the complete 48 cells and precision/recall remain in the unchanged evidence ledger and archive.}
\label{tab:compact-directional-contrasts}
\end{table}

\paragraph{Prefix dependence.}
With fresh noise on the signed pair, the final panel changes only the first two CFG values to $b\in\{1,1.5\}$ and compares Local, Full and Residual within each prefix. Midpoint/final CFG stays 1.5; $d,P,r$ are recomputed at each prefix-specific state. For a metric $M$, the interaction is $D_a^M=[M(L,1.5)-M(a,1.5)]-[M(L,1)-M(a,1)]$. Full and Residual interaction gates separately require $D^{\rm FID}\geq.05$ and $D^{\rm DINO}>0$ per model; a third gate tests high-prefix Residual/Local. All pass. Full's mean FID/DINO interaction is 1.190/11.674; Residual's is .377/5.073. Its checkpoint-specific interactions are .433/.321 FID and 5.982/4.164 DINO.

Table~\ref{tab:compact-directional-contrasts} summarizes the main contrasts. The low-prefix failure remains central: Full has mixed FID effects and worsens both KID and DINO on both models at $b=1$. Residual improves distances but trades precision for recall: changes are $-.01090/+.04420$ at $b=1$ and $-.00595/+.05135$ at $b=1.5$. Recomputing the correction at each state does not separate state sensitivity from changes in that correction, and this nonlinear metric interaction is not a fraction of Full's benefit explained.

\paragraph{Computation and limits.}
Native Local/Full use eight backbone calls. Standalone Residual uses twelve: four in the prefix, six midpoint calls to obtain three guided velocities, and two in the suffix. Shared experimental computation totals 2,160,000 sample-level forwards across 480,000 images; 3,200 smoke forwards and feature extraction are separate. The independent audit recomputes FID/DINO from cached feature arrays for all 48 cells (maximum discrepancies $7.36\times10^{-7}$/$5.53\times10^{-7}$), but does not independently re-extract features or implement KID/precision/recall. Norms, signs, projector ranks and saved intervention traces also pass their checks. The evidence identifies a useful component outside two specified local directions; it does not identify true-flow error, a universal selection principle, or a sampler superior at equal computation.

\paragraph{Gate minima and complete signed doses.}
The next two tables re-tabulate the existing frozen reports, adding no experiments. The portable renderer \path{repro/build_mechanism_detail_20260925.py} checks all 48 reports against their archived hashes and ledger, recomputes each gate at its original statistical unit, and records the inputs and unrounded margins in \path{evidence/revision_20260925/mechanism_detail.json}. This verifies report arithmetic, not independent metric extraction.

% Generated by repro/build_mechanism_detail_20260925.py from frozen reports.
\begin{table}[htbp]
\centering\small\setlength{\tabcolsep}{3.5pt}
\begin{tabular}{llcrrr}
\toprule
Panel & Contrast / gate & Units & Min. FID & Min. DINO & FID margin \\
\midrule
Projection & Full / Local (prereq.) & 4 pairs & +1.5006 & +2.7743 & -- \\
 & Residual / Local & 4 pairs & +2.6896 & +15.0465 & +2.6608 \\
 & Full / Projected & 4 pairs & +0.6452 & +8.9053 & +0.5967 \\
\addlinespace[2pt]
Equal norm & Residual / Local & 2 models & +2.3228 & +15.4099 & +2.2728 \\
 & Residual / scaled Full & 2 models & +0.9694 & +9.1436 & +0.9194 \\
 & Residual / Random & 2 models & +2.2298 & +14.4043 & +2.1798 \\
\addlinespace[2pt]
Signed & $+r$ / $-r$ & 2 models & +6.8445 & +76.0635 & +6.7945 \\
 & $+r$ / Local & 2 models & +2.2126 & +15.7994 & +2.1626 \\
\addlinespace[2pt]
Prefix & Full interaction & 2 models & +1.1389 & +10.7096 & +1.0889 \\
 & Residual interaction & 2 models & +0.3206 & +4.1635 & +0.2706 \\
 & Residual / Local, $b=1.5$ & 2 models & +2.2171 & +16.7077 & +2.1671 \\
\bottomrule
\end{tabular}
\caption{\textbf{Weakest observed unit for every frozen directional gate.} Gains are reference minus treatment; prefix interactions are high-prefix minus low-prefix gains. Min.~FID/DINO range over the stated units, which may attain their minima on different units. The final column is the minimum \emph{checkpoint-mean} FID gain minus the frozen $.05$ threshold. Projection has two repeats per model, whereas all other panels have one repeat per model. The prerequisite requires positivity only. All shown minima and margins pass; these are descriptive gates, not significance tests.}
\label{tab:revised-mechanism-gate-minima}
\end{table}

\begin{table}[htbp]
\centering\small\setlength{\tabcolsep}{5pt}
\begin{tabular}{crrrrrr}
\toprule
 & \multicolumn{3}{c}{Checkpoint 700057} & \multicolumn{3}{c}{Checkpoint 700067} \\
\cmidrule(lr){2-4}\cmidrule(lr){5-7}
Dose $\alpha$ & FID & KID $\times10^3$ & DINO & FID & KID $\times10^3$ & DINO \\
\midrule
$-1$ & 19.3259 & 8.0277 & 214.1753 & 19.6556 & 7.9593 & 221.1238 \\
$-0.5$ & 16.4837 & 6.9552 & 176.1563 & 16.5493 & 6.7097 & 180.4755 \\
0 (Local) & 14.6940 & 6.2627 & 153.9113 & 14.5998 & 5.9022 & 156.5978 \\
$+0.5$ & 13.4437 & 5.5216 & 142.4353 & 13.2309 & 5.0992 & 144.1813 \\
$+1$ & 12.4814 & 4.6672 & 138.1119 & 12.1843 & 4.2334 & 139.3422 \\
\bottomrule
\end{tabular}
\caption{\textbf{All five frozen signed doses on both existing models.} Each checkpoint--dose combination uses one 10k sample set at generation seed 202619261 with the stated midpoint velocity $v_0+\alpha r$; all other settings match the signed panel. Lower is better for all three distances. Their decline across the five doses is observed on each model; monotonicity was not a selection rule or frozen success criterion.}
\label{tab:revised-signed-doses}
\end{table}

Projection Full/Projected KID gains ($\times10^3$) are $-0.0146$ and $-0.0772$ on checkpoint 300007, and $+0.4477$ and $+0.4258$ on checkpoint 300017. Values follow generation seeds 202619241 and 202619242. Only two of four improve KID; KID is not part of the FID/DINO gates.

% Allow the following appendix to share the page after the final tables.

% Compact view generated editorially from unchanged evidence/blind_nfe_20260925/numeric_ledger.json.
% Deduplication is by (cohort,NFE,candidate_id), never by rounded metric values.
% All 30 role means, 34 contrasts and 36 raw confirmation cells remain archived.
\section{Validation of frozen candidate sets on new checkpoints}
\label{app:blind-nfe-20260925}

\subsection{Frozen candidate sets and selection procedure}
Two new CIFAR training seeds, 700081 and 700087, test the inherited horizon rule and target calibration within candidate sets frozen on older models. They use the same architecture and recipe as the source study, 250 epochs and 48,750 optimizer steps, with the final EMA checkpoint specified in advance as the evaluation checkpoint. They remain separate from the earlier seven-checkpoint panels; three NFE settings and two sampling repeats do not turn these into six or twelve training replications. We do not form a pooled nine-checkpoint average under different protocols.

\paragraph{Frozen candidate sets.}
Old development evaluates two historical checkpoints (300007/300017), NFE~2/4/8 and nine candidates with 2k images per cell: 108,000 images. Mixed chooses Local plus two of four Start and two of four Horizon candidates. The subset is selected by minimizing mean regret relative to the finite nine-candidate minimum over $\binom{4}{2}^2=36$ possibilities, separately for NFE~2 and pooled NFE~4/8 contexts; exact ties follow a fixed lexicographic rule. The resulting candidate sets are sealed before target outputs: Local/$s_1,s_2,h_1,h_2$ at NFE~2 and Local/$s_1,s_4,h_1,h_4$ at NFE~4/8. All three NFEs occur in development; this is not unseen-NFE extrapolation.

Each new model--NFE context evaluates the nine-candidate union with 5k images per candidate. Horizon-only, Start-only and Mixed each observe five candidates including Local, so each has a nominal 25,000-image target calibration budget. The physical union costs 45,000 images. Mixed additionally uses the historical development stage above: only the target budgets are equal. Minimum calibration FID determines the target choice using the frozen candidate-order tie rule. Each choice is sealed before two paired 10k confirmation repeats. No confirmation metric refits the subset, predictor, rule or threshold.

\paragraph{Post-hoc allocation diagnostic.}
Recomputation of the frozen development records finds zero-regret ties for 36/36 subsets at NFE~2 and 9/36 at NFE~4/8. Every minimizing NFE~4/8 subset contains $s_4$ and $h_4$; the remaining Start and Horizon slots follow the tie rule. At NFE~4 these two settings are the historical 25-candidate winners. The historical and new-seed comparisons use different cohorts and evaluation protocols, so their preference reversal cannot be attributed to candidate count alone. A separate post-hoc check of Local/$s_4,h_4$ attains the finite development minimum in all six old contexts and reproduces all six target choices from their saved 5k calibration scores. This three-candidate set was not prospectively enrolled: the check establishes neither a new confirmed strategy nor a lower-budget advantage. The frozen confirmations validate the tested five-candidate procedure and its additional Horizon choice at NFE~4; they do not isolate a benefit of learning the allocation. The portable script \path{repro/check_candidate_allocation_20260925.py} recomputes this diagnostic directly from archived reports, policy and selections, with input hashes in \path{evidence/revision_20260925/allocation_diagnostic.json}. It performs no new generation, selection refit or gate change.

\paragraph{Candidate domain.}
The uniform state grid has $\delta=1/\mathrm{NFE}$; only the midpoint query changes. Local uses $(\tau,h)=(.5,\delta)$. At NFE~4/8, Start uses $\tau=.5+\delta\{- .1,.1,.2,.3\}$ with $h=\delta$; Horizon uses $h=\delta\{.8,1.2,1.5,.4895833333/.25\}$ with $\tau=.5$. At NFE~2, Start uses $\tau\in\{.30,.35,.40,.45\}$ and $h=.5$; Horizon uses $h\in\{.10,.20,.30,.40\}$ and $\tau=.5$. Candidate identifiers $s_1,\ldots,s_4$ and $h_1,\ldots,h_4$ follow the listed order for each NFE; they are distinct from sampler update indices. All intervals satisfy $0\leq\tau<\tau+h\leq1$. The fixed ratio $h/\delta=1.2$ is infeasible at the NFE~2 final query, so fixed Horizon falls back to Local; this is not a nontrivial success. Thus the Horizon candidate sets at NFE~4/8 include both a shorter horizon ($.8\delta$) and longer horizons; at NFE~2, the Horizon candidate set includes shorter horizons and Local only. These are CFG-1.5 midpoint candidates, and independent confirmation covers selected interventions, not every candidate.

\paragraph{Evaluation and gates.}
CFG is 1.5 with two backbone branches per query. Calibration seed 202619282 and confirmation seeds 202619283/284 are distinct from old development. Arms within a context share saved initial noise and labels; exact selected interventions are generated once, retaining every role alias. All metrics use the same rounded uint8 images and frozen 10k real reference. Calibration computes Inception FID; confirmation computes FID/KID and DINOv2 distance, precision and recall. Absolute FIDs are not compared across different sample-size/reference protocols.

For each model and NFE, the descriptive distance gate requires mean FID gain $\geq.05$, positive FID gain in both repeats and positive mean DINO-distance gain. Precision and recall drops exceeding .01 absolute are separate flags. These are descriptive conditions, not training-population significance tests; passing the distance gate does not imply uniform quality improvement.

\subsection{Unique interventions and selected roles}
Table~\ref{tab:compact-blind-roles} maps each strategy to its selected candidate. Here $s_i$ and $h_i$ denote candidate identifiers, not sampler update indices.
\begin{table}[htbp]
\centering\small
\begin{tabular}{lrrrrl}
\toprule
Seed & NFE & Fixed H & Cal. H & Cal. Start & Mixed \\
\midrule
700081 & 2 & Local$^*$ & Local & Local & Local \\
700081 & 4 & $h_2$ & $h_4$ & $s_4$ & $h_4$ \\
700081 & 8 & $h_2$ & $h_4$ & $s_4$ & $s_4$ \\
700087 & 2 & Local$^*$ & Local & Local & Local \\
700087 & 4 & $h_2$ & $h_4$ & $s_4$ & $h_4$ \\
700087 & 8 & $h_2$ & $h_4$ & $s_4$ & $s_4$ \\
\bottomrule
\end{tabular}
\caption{All selected roles; Local is always the shared reference. $^*$Fixed extension is infeasible at NFE~2. At NFE~4 Mixed equals calibrated Horizon; at NFE~8 it equals calibrated Start. These identities explain the repeated values in the main table without counting duplicate generations.}
\label{tab:compact-blind-roles}
\end{table}

\begin{table}[htbp]
\centering\footnotesize\setlength{\tabcolsep}{3.6pt}
\begin{tabular}{lrlrrrrr}
\toprule
Seed & NFE & Arm & FID $\downarrow$ & KID$\times10^3\downarrow$ & DINO $\downarrow$ & Prec. $\uparrow$ & Rec. $\uparrow$ \\
\midrule
700081 & 2 & Local & 15.1220 & 6.2885 & 167.175 & 0.7097 & 0.3176 \\
700081 & 4 & Local & 14.1230 & 5.6144 & 145.831 & 0.7145 & 0.3684 \\
700081 & 4 & $h_2$ & 13.6702 & 5.2822 & 142.588 & 0.7109 & 0.3803 \\
700081 & 4 & $h_4$ & 12.6322 & 4.5026 & 139.306 & 0.6953 & 0.4072 \\
700081 & 4 & $s_4$ & 12.8443 & 3.9415 & 158.941 & 0.6902 & 0.3768 \\
700081 & 8 & Local & 13.0614 & 4.7497 & 137.521 & 0.7184 & 0.3905 \\
700081 & 8 & $h_2$ & 13.0086 & 4.7037 & 137.298 & 0.7157 & 0.3933 \\
700081 & 8 & $h_4$ & 12.7713 & 4.5180 & 136.424 & 0.7110 & 0.3984 \\
700081 & 8 & $s_4$ & 12.6796 & 4.1863 & 140.931 & 0.7078 & 0.3919 \\
700087 & 2 & Local & 15.0750 & 6.3345 & 172.046 & 0.7120 & 0.3236 \\
700087 & 4 & Local & 14.3259 & 5.8942 & 152.479 & 0.7136 & 0.3703 \\
700087 & 4 & $h_2$ & 13.8484 & 5.5501 & 149.747 & 0.7109 & 0.3785 \\
700087 & 4 & $h_4$ & 12.6479 & 4.7497 & 146.675 & 0.6919 & 0.4083 \\
700087 & 4 & $s_4$ & 12.9120 & 4.0774 & 166.003 & 0.6859 & 0.3795 \\
700087 & 8 & Local & 13.2883 & 5.0787 & 142.952 & 0.7145 & 0.3975 \\
700087 & 8 & $h_2$ & 13.2338 & 5.0315 & 142.789 & 0.7122 & 0.3952 \\
700087 & 8 & $h_4$ & 12.9965 & 4.8419 & 142.197 & 0.7091 & 0.3995 \\
700087 & 8 & $s_4$ & 12.9888 & 4.6119 & 145.987 & 0.7083 & 0.3989 \\
\bottomrule
\end{tabular}
\caption{Every unique selected intervention, averaging its two paired 10k repeats. There are 18 unique model--NFE--candidate combinations and 36 confirmation cells. The original 30 role means are recoverable using Table~\ref{tab:compact-blind-roles}. Precision and recall use DINOv2 features; all KID estimator SDs and F-scores remain in the unchanged numeric ledger. Rows are deduplicated by candidate identity, not by similar scores.}
\label{tab:compact-blind-unique}
\end{table}
\FloatBarrier

\subsection{Both repeats and the important failures}
Table~\ref{tab:compact-blind-unique} reports the unique selected interventions; Table~\ref{tab:compact-blind-contrasts} reports their key paired contrasts.
\begin{table}[htbp]
\centering\scriptsize\setlength{\tabcolsep}{3.5pt}
\begin{tabular}{lrlrrrrc}
\toprule
Seed & NFE & Treatment / reference & $\Delta$FID r1 & $\Delta$FID r2 & KID gain$\times10^3$ & DINO gain & Gate \\
\midrule
700081 & 4 & Fixed H / Local & +0.4531 & +0.4526 & +0.3322 & +3.244 & Pass \\
700081 & 4 & Cal. H / Local & +1.5073 & +1.4743 & +1.1117 & +6.526 & Pass \\
700081 & 4 & Cal. H / Start & +0.1650 & +0.2591 & -0.5612 & +19.636 & Pass \\
700081 & 8 & Fixed H / Local & +0.0512 & +0.0544 & +0.0459 & +0.223 & Pass \\
700081 & 8 & Cal. H / Start & -0.1095 & -0.0738 & -0.3317 & +4.507 & Fail \\
700087 & 4 & Fixed H / Local & +0.4771 & +0.4779 & +0.3441 & +2.733 & Pass \\
700087 & 4 & Cal. H / Local & +1.6281 & +1.7279 & +1.1445 & +5.804 & Pass \\
700087 & 4 & Cal. H / Start & +0.2207 & +0.3075 & -0.6722 & +19.328 & Pass \\
700087 & 8 & Fixed H / Local & +0.0495 & +0.0596 & +0.0473 & +0.164 & Pass \\
700087 & 8 & Cal. H / Start & -0.0208 & +0.0054 & -0.2300 & +3.790 & Fail \\
\bottomrule
\end{tabular}
\caption{Key paired contrasts retain both FID repeat gains and mean KID/DINO gains (reference minus treatment; positive is favorable). Gates test FID/DINO only. All 34 registered contrasts, including exact-zero identities and all precision/recall flags, remain in the ledger. NFE~2 contrasts are zero because every calibrated strategy selects Local.}
\label{tab:compact-blind-contrasts}
\end{table}

At NFE~4, both new models select Horizon, with mean FID gains over calibrated Start of \BlindFourStartMarginA{} and \BlindFourStartMarginB{}, positive in all four paired repeats. DINO also improves. \textbf{KID is higher than with Start on both models.} Relative to Local, calibrated Horizon lowers DINO precision by .0192/.0216, exceeding the .01 flag, while increasing recall by .0388/.0381. This establishes an additional useful FID/DINO choice in these configurations, not metric-uniform superiority.

At NFE~8, Mixed selects Start on both checkpoints and offers \textbf{no additional Horizon selection benefit}. Fixed Horizon still improves Local by \BlindEightFixedGainA{}/\BlindEightFixedGainB{} mean FID with small positive mean DINO gains. Calibrated Horizon has worse mean FID than Start by .0916/.0077, although its DINO distance is better; the second model's FID comparison changes sign across repeats. The FID-selected Start also worsens DINO distance relative to Local by 3.4101 and 3.0345 on the two models. At NFE~2, all calibrated strategies select Local and provide no nontrivial benefit. Target rollout calibration remains necessary for this procedure; no zero-rollout predictor or new-architecture extrapolation is claimed.

\subsection{Post-hoc sensitivity to calibration sample budget}
\label{app:budget-sensitivity}
We reuse the stored 5k-per-candidate Inception features to ask whether fewer samples would recover each frozen strategy's original choice. The existing choices and confirmation outcomes were already known. Before scoring the lower-budget selections, we seal budgets of 500/1k/2k and 20 class-balanced, paired, nested subsets of the same pool; candidates, FID evaluator, real reference and tie rules remain fixed. The 54 full-pool FIDs and all 18 strategy choices are reproduced before subsampling. This is a retrospective sensitivity analysis with two training units, no new images and no new feature extraction.

% Generated from archived summary.json; see provenance.json.
\begin{table}[htbp]
\centering\small
\setlength{\tabcolsep}{5pt}
\begin{tabular}{rlrrrrr}
\toprule
 & & \multicolumn{3}{c}{Agreement with 5k choice} & \multicolumn{2}{c}{Confirmation FID difference}\\
NFE & Seed & 500 & 1k & 2k & 1k & 2k\\
\midrule
4 & 700081 & 19/20 & 20/20 & 20/20 & +0.00000 & +0.00000\\
4 & 700087 & 20/20 & 20/20 & 20/20 & +0.00000 & +0.00000\\
8 & 700081 & 6/20 & 10/20 & 14/20 & +0.04582 & +0.02749\\
8 & 700087 & 4/20 & 6/20 & 9/20 & +0.00539 & +0.00423\\
\bottomrule
\end{tabular}
\caption{Post-hoc sample-budget sensitivity of the frozen five-candidate Mixed strategy. Budgets are samples per candidate (nominal strategy totals: 2.5k, 5k and 10k). Agreement counts reuse 20 subsets of one stored pool. FID differences compare the selected candidate with the original 5k choice on the existing two confirmation repeats, and are then averaged across subset selections; positive is worse. Both 1k/2k columns have complete confirmation coverage. They are not oracle regret or 20 independent confirmations.}
\label{tab:budget-sensitivity}
\end{table}

At NFE~4, Mixed recovers its original Horizon choice on both models in all 20 subsets at 1k and 2k per candidate. Start-only agreement at 1k is 9/20 and 20/20, becoming 20/20 on both at 2k. At NFE~8, Mixed remains sensitive to sample budget, as Table~\ref{tab:budget-sensitivity} shows. Its 1k/2k choices also worsen mean confirmation KID while improving DINO distance relative to the original 5k choice; no secondary metric changes selection. At NFE~2 every strategy retains Local at every tested budget. All three strategies' rows and metric differences are archived.

Confirmation differences use the existing two independent generation repeats; reusing them for 20 subset selections does not create 20 confirmations. At 500 samples, some selected candidates were never independently confirmed: Mixed coverage is 19/20 and 20/20 at NFE~4, and 17/20 and 19/20 at NFE~8, in seed order 700081/700087. The corresponding incomplete overall FID differences remain undefined. All lower budgets are below the 2048-dimensional feature size: sample-size-dependent FID bias prevents interpreting absolute scores across budgets as a quality curve. These dependent subsets support stability within the observed pools, not a general minimum sample budget, oracle-regret bound or zero-rollout rule. The nominal 1k strategy budget is 5k images rather than 25k, but this analysis measures no online generation-time savings and does not change the original confirmed procedure or its costs.

\paragraph{Cost, execution and audit.}
The study adds 738,000 images: 108,000 old development, 270,000 target calibration and 360,000 confirmation, in 144 generation/evaluation cells and 7,528,000 sample-level backbone forwards. Training, timing probes, smoke calls and feature extraction are separate. Together with the previously scoped 2,830,000-image supplementary subtotal, this gave 3,568,000 images. The later public GS/GSH study adds 244,000 (Appendix~\ref{app:public-grid-increment}), bringing the current scoped subtotal to 3,812,000; neither subtotal is a full-project total.

The second training job moved from physical GPUs 0/2 to 5/6 through a checkpoint resume. World size stayed two, local batch 128, global batch 256, with unchanged classwise matching pools, frozen source snapshot and final-EMA criterion. No four-GPU training variant was deployed; uninterrupted data/RNG bitwise identity is not asserted. Device authorization and adapters are recorded outside the unchanged scientific protocol. The evidence archive retains 144 evaluation reports and their 144 corresponding generation records, 191 sealed source files, selection chronology, paired hashes and final training certificates. Its CPU audit checks file bindings and arithmetic, not independent feature re-extraction. The complete role/repeat tables remain in the archived full manuscript.
\FloatBarrier

% Compact boundary record. Values are retained from prediction_20260922.tex,
% temporal_followup.tex, guidance_evidence.tex and appendix.tex; full per-cell
% results remain in the unchanged evidence ledgers and archived full manuscript.
\section{Prediction and transfer boundaries}

\subsection{Prospective diagnostic prediction did not meet its gate}
\label{app:prediction-20260922}
The frozen predictor tests whether midpoint short/long-velocity disagreement improves prediction of the eventual Local-minus-Horizon FID gain. Development consists of twelve historical contrasts on six checkpoints (three MAC-paired and three semidiscrete), at prefix CFG 1/1.5. Test uses the \emph{same weights} at new prefixes .75/1.25, not unseen checkpoints. Within each Local/Horizon comparison, the grid and suffix CFG remain fixed; only the midpoint horizon changes from $h=.25$ to $.50$. The diagnostic is
\begin{equation}
z=\log\!\left(1+\frac1{1024}\sum_i
\frac{\|u_l^{(i)}-u_s^{(i)}\|_2^2}{\|u_s^{(i)}\|_2^2+10^{-12}}\right).
\end{equation}
It adds one shared feature to a recipe--prefix ridge regression (penalty $10^{-3}$ on standardized non-intercept terms). Baselines are a constant, the recipe--prefix model and each checkpoint's line through its two historical gains. Coefficients, development standardization and all test predictions were sealed before generating 240,000 new endpoint images. The 24,576 midpoint probes and 196,608 probe backbone calls are additional work and include prefix rollouts.

The conjunctive success gate requires at least 10\% lower overall MAE than \emph{both} recipe--prefix and checkpoint-line baselines, plus lower MAE than recipe--prefix within each recipe and each prefix. The diagnostic MAE is \textbf{0.303023}, versus \textbf{0.242689} for the checkpoint line; the gate fails. Constant and recipe--prefix MAEs are .97444 and .37212. Improvements over recipe--prefix and favorable extrapolation behavior do not overturn the stronger baseline's overall advantage. All six gains at prefix .75 are negative; all six at prefix 1.25 are positive. Secondary sign accuracy (11/12) and decision regret (.00767 versus .01572 for the checkpoint line) remain secondary. No feature/model/threshold rescue is performed on these outcomes. The audit passes, but prediction success does not; this result does not justify replacing target endpoint calibration or predicting unseen-checkpoint gains.

\subsection{Native iMF and public-model prefix checks}
\label{app:imf-native}
\label{app:imf-position-followup}
The official iMF-XL/2 implementation reproduces its native 50k one/two-query endpoints (FID 1.7238/1.5443, IS 279.93/288.88). Its interface ignores the explicit start-time argument, so Start is exactly Local and Joint exactly Horizon in that adapter. A separate four-query, midpoint-$h=.30$ pilot selects native guidance 4 from $\{1,2,4,8\}$ before testing. Its paired FID gains are .0227/.0159, mean .0193, with approximately 2.1\% lower IS. It fails the .05 mean-gain continuation threshold.

The subsequently frozen position--horizon panel uses the same checkpoint, four uniform updates, native guidance 4 and guidance interval $[.36,.64]$. It evaluates six interventions plus Local on two fresh 10k seeds. The descriptive gate requires both FID gains positive, mean gain $\geq.05$, and mean IS retention $\geq.95$. \textbf{None passes}; the best observed mean gain is only .0040, an exploratory maximum. Table~\ref{tab:imf-position-panel} reports every intervention. These repeats remain one public training checkpoint.

% Generated from SHA-256-verified JSON.
\begin{table}[ht]
\centering
\small
\caption{Native iMF-XL/2 fixed position--horizon panel. Two new generation repeats, each 10k, at native guidance 4. FID gains are paired Local minus intervention; positive is better. IS retention is the mean ratio to that local arm. This is one checkpoint; the largest panel gain is an exploratory maximum.}
\label{tab:imf-position-panel}
\begin{tabular}{rrrrrr}
\toprule
$t_k$ & $h_k$ & FID gain r1 & FID gain r2 & Mean FID gain & IS retention \\
\midrule
0.00 & 0.30 & -0.1171 & -0.0992 & -0.1081 & 0.9817 \\
0.00 & 0.50 & -2.8806 & -2.9079 & -2.8942 & 0.8698 \\
0.25 & 0.30 & -0.0518 & -0.0280 & -0.0399 & 0.9947 \\
0.25 & 0.50 & -2.2329 & -2.1834 & -2.2082 & 0.8608 \\
0.50 & 0.30 & +0.0016 & +0.0064 & +0.0040 & 0.9790 \\
0.50 & 0.50 & -1.1703 & -1.1361 & -1.1532 & 0.8744 \\
\bottomrule
\end{tabular}
\end{table}

\paragraph{Public-model guidance history.}
\label{app:public-prefix-followup}
Released MeanFlow-B/2 uses four uniform updates, CFG $(p,p,1.5,1.5)$ with $p\in\{1,1.5\}$, and midpoint $h\in\{.25,.30\}$. Four paired 10k arms per seed share noise and labels; Local/Horizon have identical entering states within prefix. All arms execute eight backbone calls. With $G_p$ denoting Local-minus-Horizon FID, the two interactions $G_{1.5}-G_1$ are \textbf{+.023485 and -.014275}. The registered directional condition requires both positive and therefore fails. Mean interaction .004605 does not supply stable transfer of the CIFAR prefix direction. Learned guidance, architecture and the different long horizon remain confounded explanations, not resolved mechanisms.

\subsection{Earlier surrogate and reference diagnostics}
\label{app:larger-spaces}
These earlier allocation studies use a low-dimensional rollout score $L$ to
choose sets $S$ of discrete horizon changes, allowing at most one change per
position. Unary selection uses $c_i=L(\{i\})-L(\varnothing)$; the pairwise
model also uses
\[
q_{ij}=L(\{i,j\})-L(\{i\})-L(\{j\})+L(\varnothing),\qquad
E(S)=\sum_{i\in S}c_i+\sum_{\substack{i,j\in S\\i<j}}q_{ij}.
\]
CIFAR uses the frozen Inception network's 64-dimensional output for FID;
ImageNet projects official 2048-dimensional Inception features onto the top
64 eigenvectors of the fixed real covariance without whitening. These scores
are distinct from the final 2048-dimensional FID. Calibration measures all
singles and compatible pairs. NFE~8 permits at most two actions, making this model
exact for the centered development score $L(S)-L(\varnothing)$ on its feasible sets; NFE~16 permits
three, leaving unmeasured three-way terms. Calibration uses 2,000 samples per
configuration at CIFAR NFE~8, 2,048 at CIFAR NFE~16 and 512 at ImageNet NFE~16;
selected schedules use disjoint evaluation noise.

At NFE~8, pairwise selection improves on unary selection in its earlier CIFAR panel; exact and greedy pairwise schedules coincide. This does not validate the surrogate at higher NFE. CIFAR NFE~16 evaluates 991 baseline/single/pair configurations for 45 actions and 13,276 feasible three-action schedules. Exact and greedy again coincide, so its predeclared greedy-comparison gate stops the study before new 50k evaluation.

ImageNet NFE~16 directly exposes the ranking failure: Local has FID 7.6307; unary has energy $-.6702$ and FID \textbf{7.2473}; pairwise has better surrogate energy $-.7388$ but worse FID \textbf{7.3102}. Lower surrogate energy therefore does not imply lower rollout FID. Higher-order effects, finite calibration noise and objective mismatch are possible explanations, whose contributions are unresolved.

\label{app:negative}
A separate conditional-allocation pilot uses eight semantic groups and five paired 10k validation arms. Conditional pairwise FID 7.2914 improves universal 7.3108 by only .0194, below the frozen .05 threshold; conditional unary is 7.3059. Exact and greedy coincide in all eight groups. The study stops at its 10k stage rather than expanding the search.

\paragraph{Numerical and analytic references.}
\label{app:reference}
The model-defined reference study uses 1,024 states from the historical
seed-42 checkpoint. From each state it integrates the next $\delta=.25$
interval with 16 native local or instantaneous-Heun substeps, obtaining the
reference average velocity $(x_{\rm ref}-x)/\delta$. For each sample it divides
the mean squared velocity error by the reference's mean squared coordinates,
with denominator floored at $10^{-12}$, then averages over samples.
At $t=.5$, these normalized errors to the native local reference increase with $h=.25,.30,.375,.50$: .000789/.001268/.003020/.008595. The corresponding instantaneous-Heun errors are .000882/.001370/.003135/.008725. Thus the horizon that improves endpoint FID is not more accurate against these model-defined local references. They are not ground-truth transport paths.

\label{app:oracle-boundary}
The separate analytic two-dimensional diagnostic uses eight Gaussian components
on a ring and an analytic conditional-expectation flow field. Two small
regressors learn RK4 oracle interval averages, not the MeanFlow objective,
and are evaluated after 100, 500 and 2,000 updates: six model--training-stage
states, not six independently trained networks. At each state, the six fixed
interventions use $h\in\{.30,.50\}$ at one of the first three positions of a
four-step grid. Local-MSE selection compares velocity predictions with the
oracle average over the physical $.25$ update on 1,024 probe inputs. Endpoint
evaluation uses separate rollout noise, 4,096 samples and an independent target
draw, measuring squared sliced-Wasserstein distance over 64 directions.
Across the 36 interventions, local-MSE selection predicts gain signs correctly for 30/36, below the always-no-gain baseline's 33/36. It chooses harmful interventions at both 500-update checkpoints and misses available gains at both final checkpoints, despite converged numerical references (relative RMSE $5.16\times10^{-8}$). This supports retaining endpoint validation, not a general impossibility claim for future selectors. Every intervention, prediction and stopping decision is preserved in the evidence archive.
\FloatBarrier

% Detailed archive navigation and accounting are in repro/EVIDENCE_INDEX.md.
\end{document}